\documentclass[10pt,journal,compsoc]{IEEEtran}

\usepackage[numbers,sort&compress]{natbib}
\usepackage[pdftex]{graphicx}
\usepackage[cmex10]{amsmath}
\usepackage{hyperref}
\usepackage{url}
\usepackage{paralist}
\usepackage{multirow}
\usepackage{color}
\usepackage{xcolor}
\usepackage{booktabs}
\usepackage{commath}
\usepackage{latexsym}

\newcommand{\eg}{\emph{e.g.},}
\newcommand{\ie}{\emph{i.e.},}

\newcommand{\define}{\:{\scriptstyle\stackrel{\triangle}{=}}\:}
\newcommand{\degr}{$^\circ$}
\newcommand{\rot}{\textrm{rot}}

\definecolor{dataset1set1}{RGB}{0,0,144}
\definecolor{dataset1set2}{RGB}{0,0,255}
\definecolor{dataset1set3}{RGB}{0,112,255}
\definecolor{dataset1set4}{RGB}{0,223,255}
\definecolor{dataset1set5}{RGB}{80,255,175}
\definecolor{dataset1set6}{RGB}{191,255,64}
\definecolor{dataset1set7}{RGB}{255,207,0}
\definecolor{dataset1set8}{RGB}{255,96,0}
\definecolor{dataset1set9}{RGB}{239,0,0}
\definecolor{dataset1set10}{RGB}{127,0,0}

\definecolor{dataset2set1}{RGB}{0,0,143}
\definecolor{dataset2set2}{RGB}{0,143,255}
\definecolor{dataset2set3}{RGB}{143,255,112}
\definecolor{dataset2set4}{RGB}{255,112,0}
\definecolor{dataset2set5}{RGB}{127,0,0}

\newsavebox{\boxaa}\sbox{\boxaa}{\textcolor{dataset1set1}{\rule{6pt}{6pt}}}
\newsavebox{\boxab}\sbox{\boxab}{\textcolor{dataset1set2}{\rule{6pt}{6pt}}}
\newsavebox{\boxac}\sbox{\boxac}{\textcolor{dataset1set3}{\rule{6pt}{6pt}}}
\newsavebox{\boxad}\sbox{\boxad}{\textcolor{dataset1set4}{\rule{6pt}{6pt}}}
\newsavebox{\boxae}\sbox{\boxae}{\textcolor{dataset1set5}{\rule{6pt}{6pt}}}
\newsavebox{\boxaf}\sbox{\boxaf}{\textcolor{dataset1set6}{\rule{6pt}{6pt}}}
\newsavebox{\boxag}\sbox{\boxag}{\textcolor{dataset1set7}{\rule{6pt}{6pt}}}
\newsavebox{\boxah}\sbox{\boxah}{\textcolor{dataset1set8}{\rule{6pt}{6pt}}}
\newsavebox{\boxai}\sbox{\boxai}{\textcolor{dataset1set9}{\rule{6pt}{6pt}}}
\newsavebox{\boxaj}\sbox{\boxaj}{\textcolor{dataset1set10}{\rule{6pt}{6pt}}}

\newsavebox{\boxba}\sbox{\boxba}{\textcolor{dataset2set1}{\rule{6pt}{6pt}}}
\newsavebox{\boxbb}\sbox{\boxbb}{\textcolor{dataset2set2}{\rule{6pt}{6pt}}}
\newsavebox{\boxbc}\sbox{\boxbc}{\textcolor{dataset2set3}{\rule{6pt}{6pt}}}
\newsavebox{\boxbd}\sbox{\boxbd}{\textcolor{dataset2set4}{\rule{6pt}{6pt}}}
\newsavebox{\boxbe}\sbox{\boxbe}{\textcolor{dataset2set5}{\rule{6pt}{6pt}}}

\graphicspath{{images/}}
\DeclareGraphicsExtensions{.png,.jpg,.pdf}

\begin{document}

\title{Sentence Directed Video Object Codetection}

\author{Haonan~Yu,~\IEEEmembership{Student~Member,~IEEE}
        and~Jeffrey~Mark~Siskind,~\IEEEmembership{Senior~Member,~IEEE}%
\thanks{Both H.~Yu and J.~M.~Siskind are with the School of Electrical
  and Computer Engineering, Purdue University, West Lafayette, IN 47907
  USA.\protect\\
  E-mail: haonan@haonanyu.com, qobi@purdue.edu}%
\thanks{Manuscript received}}

\markboth{IEEE TRANSACTIONS ON PATTERN ANALYSIS AND MACHINE INTELLIGENCE}%
{Yu \& Siskind: Sentence Directed Video Object Codetection}

\IEEEtitleabstractindextext{%
\begin{abstract}
We tackle the problem of video object codetection by leveraging the weak
semantic constraint implied by sentences that describe the video
content.
Unlike most existing work that focuses on codetecting large objects
which are usually salient both in size and appearance, we can codetect
objects that are small or medium sized.
Our method assumes no human pose or depth information such as is
required by the most recent state-of-the-art method.
We employ weak semantic constraint on the codetection process by
pairing the video with sentences.
Although the semantic information is usually simple and weak, it can
greatly boost the performance of our codetection framework by reducing
the search space of the hypothesized object detections.
Our experiment demonstrates an average IoU score of 0.423 on a new
challenging dataset which contains 15 object classes and 150 videos with 12,509
frames in total, and an average IoU score of 0.373 on a subset of an
existing dataset, originally intended for activity recognition, which contains
5 object classes and 75 videos with 8,854 frames in total.
\end{abstract}

}

\maketitle

\IEEEdisplaynontitleabstractindextext

\IEEEpeerreviewmaketitle

\IEEEraisesectionheading{\section{Introduction}\label{sec:introduction}}

\IEEEPARstart{I}{n} this paper, we address the problem of codetecting objects with
bounding boxes from a set of videos, \emph{without} any pretrained
object detectors.
The codetection problem is typically approached by selecting one out
of many object proposals per image or frame that maximizes a combination of
the confidence scores associated with the selected proposals and the
similarity scores between proposal pairs.
While much prior work focuses on codetecting objects in still images
(\eg\ \citep{Blaschko2010, Lee2011, Rubinstein2013, Tang2014}), little prior
work \citep{Prest2012, Schulter2013, Armand2014, Srikantha2014, Ramanathan2014}
attempts to codetect objects in video.
In both lines of work, most \citep{Blaschko2010, Lee2011, Rubinstein2013,
  Tang2014, Prest2012, Armand2014} assume that the objects to be codetected are
salient, both in size and appearance, and located in the center of the field of
view.
Thus they easily ``pop out.''
As a result, prior methods succeed with a small number of object proposals in
each image or frame.
\citet{Tang2014} and \citet{Armand2014} used approximately 10 to 20 proposals
per image, while \citet{Lee2011} used 50 proposals per image.
Limiting codetection to objects in the center of the field of view allowed
\citet{Prest2012} to prune the search space by penalizing proposals in contact
with the image perimeter.
Moreover, under these constraints, the confidence score associated with
proposals is a reliable measure of salience and a good indicator of which image
regions constitute potential objects \citep{Rubinstein2013}.
In prior work, the proposal confidence dominates the overall scoring process
and the similarity measure only serves to refine the confidence.
In contrast, \citet{Srikantha2014} attempt to codetect small to medium sized
objects in video, without the above simplifying assumptions.
However, in order to search through the larger resulting object proposal
space, they avail themselves of human pose and depth information to prune the
search space.
It should also be noted that all these codetection methods, whether for images
or video, codetect only one common object at a time: different object classes
are codetected independently.

\begin{figure*}[t]
  \centering
  \resizebox{0.8\textwidth}{!}{
    \begin{tabular}{ccc}
      \includegraphics[width=0.33\textwidth]{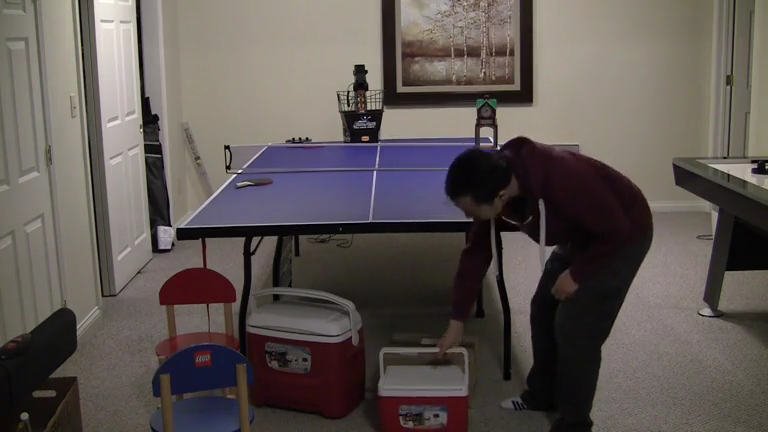}
      &\includegraphics[width=0.33\textwidth]{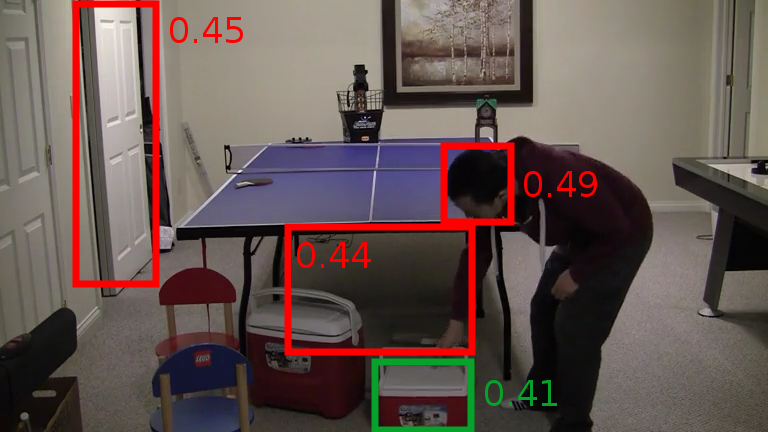}
      &\includegraphics[width=0.33\textwidth]{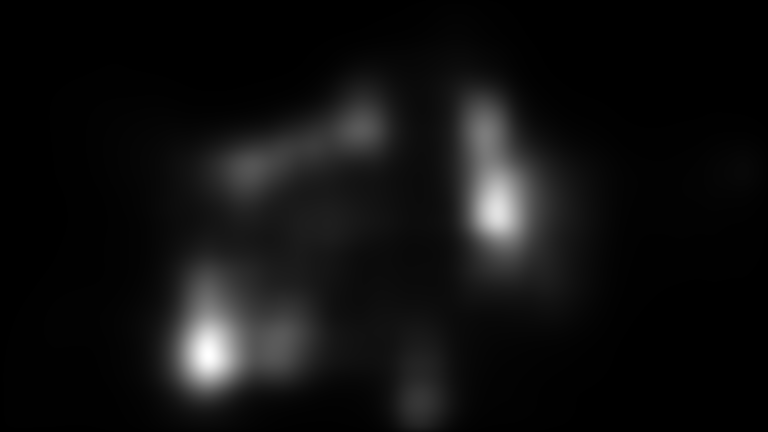}
    \end{tabular}}
  \caption{Object proposal confidence scores and saliency scores for a sample
    frame from our new dataset.
    Left: the original input video frame.
    Middle: several proposals and associated confidence scores produced by
    the method of \citet{Arbelaez2014}.
    Note that the red boxes, which do not correspond to objects, let alone
    salient ones, all have higher scores than the green box, which does denote
    a salient object.
    Right: the saliency map output by the saliency detection method of
    \citet{Jiang2015}, currently the highest ranking method on the MIT saliency
    benchmark \citep{Bylinskii2012}.
    Note that the \emph{cooler} is not highlighted as salient.
    Using these scores as part of the scoring function can drive the codetection
    process to produce undesired results.}
  \label{fig:unreliable}
\end{figure*}

The confidence score of a proposal can be a poor indicator of whether a
proposal denotes a salient object, especially when objects are occluded, the
lighting is poor, or motion blur exists (\eg\ see Figure~\ref{fig:unreliable}).
Salient objects can have low confidence score while nonsalient objects or
image regions that do not correspond to objects can have high confidence score.
Thus our scoring function does not use the confidence scores produced by the
proposal generation mechanism.
Moreover, our method does not rely on human pose and depth information, which
is not always available.
Human pose can be difficult to estimate reliably when a person is only
partially visible or is self-occluded \citep{Andriluka2014}, as is the case
with most of our videos.

We avail ourselves of a different source of constraint on the codetection
problem.
In videos depicting human interaction with objects to be codetected,
descriptions of such activity can impart \emph{weak} spatial or motion
constraint either on a single object or among multiple objects of interest.
For example, if the video depicts a ``pick up'' event, some object should have
an upward displacement during this process, which should be detectable even if
it is small.
This motion constraint will reliably differentiate the object which is being
picked up from other stationary background objects.
It is weak because it might not totally resolve the ambiguity; other image
regions might satisfy this constraint, perhaps due to noise.
Similarly, if we know object~$A$ is on the left of object~$B$, then the
detection search for object~$A$ will weakly affect the detection search for
object~$B$, and vice versa.
To this end, we extract spatio-temporal constraints from \emph{sentences} that
describe the videos and then impose these constraints on the codetection
process to find the most salient collections of objects that satisfy these
constraints.
Even though the constraints implied by a single sentence are usually
weak, when accumulated across a set of videos and sentences, they together
will greatly prune the detection search space.
We call this process \emph{sentence directed} video object codetection.
It can be viewed as the \emph{inverse} of video captioning/description
\citep{Barbu2012, Das2013, Guadarrama2013} where object evidence (detections or
other visual features) is first produced by pretrained detectors and then
sentences are generated given the object appearance and movement.

Generally speaking, we extract a set of predicates from each sentence
and formulate each predicate around a set of primitive
\emph{functions}.
The predicates may be verbs (\eg\ \textsc{carried} and \textsc{rotated}),
spatial-relation prepositions (\eg\ \textsc{toTheLeftOf} and \textsc{above}),
motion prepositions (\eg\ \textsc{awayFrom} and \textsc{towards}), or adverbs
(\eg\ \textsc{quickly} and \textsc{slowly}).
The sentential predicates are applied to the candidate object proposals as
arguments, allowing an overall predicate score to be computed that indicates
how well these candidate object proposals satisfy the sentence semantics.
We add this predicate score into the codetection framework, on top of the
original similarity score, to guide the optimization.
To the best of our knowledge, this is the first work that uses sentences to
guide generic video object codetection.
To summarize, our approach differs from the indicated prior work in the
following ways:
\begin{compactenum}[(a)]
\item Our method can codetect small or medium sized non-salient objects which
  can be located anywhere in the field of view.
\item Our method does not require or assume human pose or depth information.
\item Our method can codetect multiple objects simultaneously.
  These objects can be either moving in the foreground or
  stationary in the background.
\item Our method allows fast object movement and motion blur.
  Such is not exhibited in prior work.
\item Our method leverages sentence semantics to help codetection.
\end{compactenum}
We evaluate our approach on two different datasets.
The first is a new dataset that contains~15 distinct object classes and
150~video clips with a total of 12,509 frames.
The second is a subset of CAD-120 \citep{Koppula2013}, a dataset originally
intended for activity recognition, that contains 5~distinct object classes and
75~video clips with a total of 8,854 frames.
Our approach achieves an average IoU (Intersection-over-Union) score of 0.423
on the former and 0.373 on the latter.
It yields an average detection accuracy of 0.7 to 0.8 on the former (when the
IoU threshold is 0.4 to 0.3) and 0.5 to 0.6 on the latter (when the IoU
threshold is 0.4 to 0.3).

\section{Related Work}

Corecognition is a simpler variant of codetection \citep{Tuytelaars2010},
where the objects of interest are sufficiently prominent in the field of view
that the problem does not require object localization.
Thus corecognition operates like unsupervised clustering, using feature
extraction and the similarity measure.
Codetection \citep{Blaschko2010, Lee2011, Tang2014} additionally requires
localization, often by putting bounding boxes around the objects.
This can require combinatorial search over a large space of possible object
locations.
One way to remedy this is to limit the space of possible object locations to
those produced by an object proposal method
\citep{Alexe2010, Arbelaez2014, Zitnick2014, Cheng2014}.
These methods typically associate a confidence score with each proposal which
can be used to prune or prioritize the search.
Codetection is typically formulated as the process of selecting one proposal
per image or frame, out of the many produced by the proposal mechanism, that
maximizes the collective confidence of and similarity between the selected
proposals.
This optimization is usually performed with Belief Propagation
\citep{Pearl1982} or with nonlinear programming.
Recently, the codetection problem has been extended to video
\citep{Schulter2013, Prest2012, Armand2014, Srikantha2014, Ramanathan2014}.
Like \citet{Srikantha2014}, we codetect small and medium objects, but do
so without using human pose or a depth map.
Like \citet{Schulter2013}, we codetect both moving and stationary objects, but
do so with a larger set of object classes and a larger video corpus.
Also, like \citet{Ramanathan2014}, we use sentences to guide video
codetection, but do so for a vocabulary that goes beyond pronouns, nominals,
and names that are used to codetect only human face tracks.

Another line of work learns visual structures or models from image captions
\citep{Moringen2008, Jamieson2010a, Jamieson2010b, Luo2009, Berg2004,
  Gupta2008}, treating the input as a parallel image-text dataset.
Since this work focuses on images and not video, the sentential
captions only contain static concepts, such as the names of people or
the spatial relations between objects in the images.
In contrast, our approach models the motion and changing spatial relations that
are present only in video as described by verbs and motion prepositions in the
sentential annotation.

\section{Sentence Directed Codetection}
\label{sec:algorithm}

\begin{figure*}[t]
  \centering
  \includegraphics[width=0.8\textwidth]{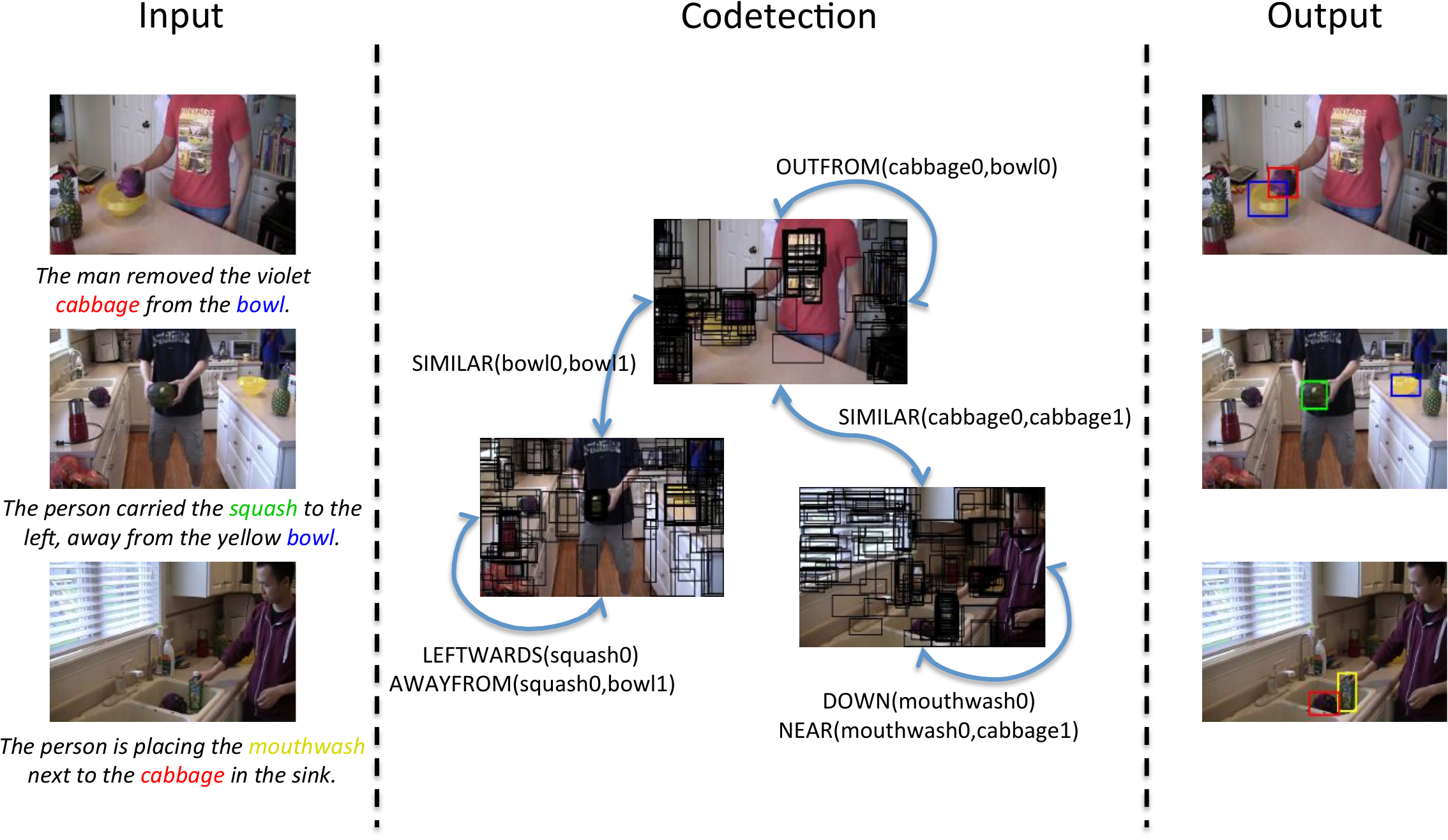}
  \caption{
    An overview of our codetection process.
    Left: input a set of videos paired with sentences.
    Middle: sentence directed codetection, where black
    bounding boxes represent object proposals.
    Right: output original videos with objects codetected.
    Note that no pretrained object detectors are used in this whole process.
    Also note how sentence semantics plays an important role in this
    process: it provides both unary scores,
    \eg\ $\textsc{leftwards}(\textit{squash0})$ and
    $\textsc{down}(\textit{mouthwash0})$, for proposal confidence, and binary
    scores, \eg\ $\textsc{outFrom}(\textit{cabbage0},\textit{bowl0})$ and
    $\textsc{near}(\textit{mouthwash0},\textit{cabbage1})$, for relating
    multiple objects in the same video.
    (Best viewed in color.)}
  \label{fig:pipeline}
\end{figure*}

Our sentence-directed codetection approach is illustrated in
Figure~\ref{fig:pipeline}.
The input is a set of videos paired with human-elicited sentences, one
sentence per video.
A collection of object-candidate generators and video-tracking methods are
applied to each video to obtain a pool of object proposals.\footnote{For
  clarity, in the remainder of this paper, we refer to object proposals for a
  single frame as object candidates, while we refer to object tubes or tracks
  across a video as object proposals.}
Object instances and predicates are extracted from the paired sentence.
Given multiple such video-sentence pairs, a graph is formed where object
instances serve as vertices and similarities between object instances and
predicates linking object instances in a sentence serve as edges.
Finally, Belief Propagation is applied to this graph to jointly infer
object codetections.

\subsection{Sentence Semantics}
\label{sec:predicate}

Our main contribution is exploiting sentence semantics to help the codetection
process.
We use a conjunction of predicates to represent (a portion of) the semantics of
a sentence.
Object instances in a sentence fill the arguments of the predicates in
that sentence.
An object instance that fills the arguments of multiple predicates is
said to be \emph{coreferenced}.
For a coreferenced object instance, only one track is codetected.
For example, a sentence like ``\textit{the person put the mouthwash
  into the sink near the cabbage}'' implies the following conjunction of
predicates:
\begin{equation*}
  \textsc{down}(\textit{mouthwash})\wedge
  \textsc{near}(\textit{mouthwash},\textit{cabbage})
\end{equation*}
In this case, \textit{mouthwash} is coreferenced by the predicates
\textsc{down} (fills the sole argument) and \textsc{near} (fills the
first argument).
Thus only one \textit{mouthwash} track will be produced, simultaneously
constrained by the two predicates (Figure~\ref{fig:pipeline}, blue track).

In principle, one could map sentences to conjunctions of our predicates using
standard semantic parsing techniques \citep{Wong2007, Clarke2010}.
However, modern semantic parsers are domain specific, and employ
machine-learning methods to train a semantic parser for a specific domain.
No existing semantic parser has been trained on our domain.
Training a new semantic parser requires a parallel corpus of sentences paired
with intended semantic representations.
Modern semantic parsers are trained with corpora like PropBank
\citep{Palmer2005} that have tens of thousands of manually annotated sentences.
Gathering such a large training corpus would be overkill for our experiments
that involve only a few hundred sentences, especially since such is not our
focus or contribution.
Thus like \citet{Lin2014}, \citet{Kong2014}, and \citet{Plummer2015}, we employ
simpler handwritten rules to fully automate the semantic parsing process for
our limited corpus.
Nothing, in principle, precludes using a machine-trained semantic parser in its
place.

Our semantic parser employs seven steps.
\begin{compactenum}
\item Spelling errors are corrected with Ispell.
\item The NLTK parser\footnote{\texttt{\url{http://www.nltk.org/}}} is used to
  obtain the POS tags for each word in the sentence.
\item POS tagging errors are corrected by a postprocessing step with a small
  set of rules (Table~\ref{tab:semantic-parser}a).
  \label{item:a}
\item Words with a specified set of POS tags\footnote{PRP\$/possessive-pronoun,
  RN/adverb, ,/comma, ./period, JJ/adjective, CC/coordinating-conjunction,
  CD/cardinal-number, DT/determiner, and JJR/adjective-comparative} are
  eliminated.
\item NLTK is used to lemmatize all nouns and verbs.
\item Synonyms are conflated by mapping phrases to a smaller set of nouns and
  verbs using a small set of rules (Table~\ref{tab:semantic-parser}b).
  \label{item:b}
\item A small set of rules map the resulting word strings to predicates
  (Table~\ref{tab:semantic-parser}c).
  \label{item:c}
\end{compactenum}
The entire process is fully automatic and implemented in less than two pages of
Python code.


\begin{table*}
  \centering
  \resizebox{\textwidth}{!}{\begin{tabular}{lll}
    \begin{tabular}[t]{@{}l@{}}
      \emph{towards}/IN\\
      \emph{ice chest}/NN\\
      \emph{watering can}/NN\\
      \emph{vegetable}/NN\\
      \emph{watering pot}/NN\\
      \emph{gas can}/NN\\
      \emph{poured}/VBD\\
      \emph{pineapple}/NN\\
      \emph{box}/NN\\
      \emph{blue}/JJ\\
      \emph{violet}/JJ\\
      \emph{pours}/VBZ\\
      \emph{underneath}/IN\\
      \emph{off}/IN\\
      \emph{inside}/IN\\
      \emph{place}/VB
    \end{tabular}&
    \begin{tabular}[t]{@{}l@{}}
      $\{\emph{gas can},\emph{gasoline can},\emph{gasoline tank}\}$\\
      $\{\emph{put},\emph{set},\emph{place}\}$\\
      $\{\emph{pick up},\emph{lift}\}$\\
      $\{\emph{milk},\emph{almond milk},\emph{carton}\}$\\
      $\{\emph{cooler},\emph{ice chest}\}$\\
      $\{\emph{table},\emph{tennis table},\emph{ping pong table}\}$\\
      $\{\emph{ground},\emph{driveway},\emph{floor}\}$\\
      $\{\emph{bowl},\emph{dish},\emph{plate}\}$\\
      $\{\emph{bucket},\emph{pail}\}$\\
      $\{\emph{cabbage},\emph{vegetable}\}$\\
      $\{\emph{box},\emph{cardboard box}\}$\\
      $\{\emph{person},\emph{man},\emph{boy}\}$\\
      $\{\emph{leftwards},\emph{leftward}\}$\\
      $\{\emph{out},\emph{outside}\}$\\
      $\{\emph{into},\emph{inside of},\emph{inside}\}$\\
      $\{\emph{towards},\emph{toward}\}$\\
      $\{\emph{on},\emph{onto}\}$
    \end{tabular}&
    \begin{tabular}[t]{@{}l@{}}
      \rule{0pt}{0pt}\\[-2ex]
      \begin{math}
        \left.\begin{array}{l}
          \emph{person carry}\;x\;\emph{rightwards put it near}\;y\\
          \emph{person carry}\;x\;\emph{leftwards put it near}\;y\\
          \emph{person carry}\;x\;\emph{to right near}\;y\\
          \emph{person carry}\;x\;\emph{to left near}\;y\\
          \emph{person put}\;x\;\emph{on right near}\;y\\
          \emph{person move}\;x\;\emph{from left to right near}\;y\\
          \emph{person carry}\;x\;\emph{to left put it near}\;y
        \end{array}\right\}\leadsto
        \textsc{moveHorizontal}(x)\wedge\textsc{nearEnd}(x,y)
      \end{math}\\
      \begin{math}
        \left.\begin{array}{l}
          \emph{person move}\;x\;\emph{out of}\;y\\
          \emph{person take}\;x\;\emph{out of}\;y\;\emph{put it to right}\\
          \emph{person take}\;x\;\emph{out of}\;y\\
          \emph{person take}\;x\;\emph{from}\;y\;\emph{put it on counter}
        \end{array}\right\}\leadsto
        \textsc{inStart}(x,y)\wedge\textsc{awayFrom}(x,y)
      \end{math}
    \end{tabular}\\
    \hline
    \begin{tabular}[t]{@{}l@{}}
      \emph{up}/RB\\
      \emph{down}/RB\\
      \emph{blue}/JJ\\
      \emph{drank}/VBD\\
      \emph{pours}/VBZ\\
      \emph{poured}/VBD\\
      \emph{places}/VBZ\\
      \emph{reaches}/VBZ\\
      \emph{bowl}/NN
    \end{tabular}&
    \begin{tabular}[t]{@{}l@{}}
      $\{\emph{put},\emph{set},\emph{stack},\emph{place}\}$\\
      $\{\emph{pick up},\emph{raise},\emph{lift}\}$\\
      $\{\emph{drink},\emph{take drink},\emph{pick up drink}\}$\\
      $\{\emph{take},\emph{remove}\}$\\
      $\{\emph{cup},\emph{container},\emph{mug},\emph{tin}\}$\\
      $\{\emph{cereal},\emph{cereal box},\emph{box}\}$\\
      $\{\emph{water},\emph{liquid}\}$\\
      $\{\emph{ground},\emph{floor}\}$\\
      $\{\emph{on},\emph{onto}\}$
    \end{tabular}&
    \begin{tabular}[t]{@{}l@{}}
      \rule{0pt}{0pt}\\[-2ex]
      \begin{math}
        \left.\begin{array}{l}
          \emph{person put}\;x\;\emph{into table}\\
          \emph{person put}\;x\;\emph{on table}\\
          \emph{person put down}\;x\\
          \emph{person put}\;x\;\emph{down on table}\\
          \emph{person put}\;x\;\emph{down}
        \end{array}\right\}\leadsto\textsc{moveDown}(x)
      \end{math}\\
      \begin{math}
        \left.\begin{array}{l}
          \emph{person pour}\;x\;\emph{into}\;y\\
          \emph{person pick up}\;x\;\emph{pour it into}\;y\\
          \emph{person pick up pour}\;x\;\emph{into}\;y
        \end{array}\right\}\leadsto\textsc{rotate}(x)\wedge\textsc{over}(x,y)
      \end{math}
    \end{tabular}\\
    \multicolumn{1}{c}{(a)}&\multicolumn{1}{c}{(b)}&\multicolumn{1}{c}{(c)}
  \end{tabular}}
  \caption{Sample rules from (a)~step~\ref{item:a}, (b)~step~\ref{item:b}, and
    (c)~step~\ref{item:c} of our semantic parser.
    (top)~For our new dataset.
    (bottom)~For the subset of CAD-120.}
  \label{tab:semantic-parser}
\end{table*}

The rules employed by the last step of the above process generate a \emph{weak}
semantic representation, containing only those predicates that are relevant to
our codetection process.
For example, for the phrase ``\textit{into the sink}'' in the above
sentence, it is beyond our interest to detect the object \textit{sink}.
Thus our predefined rules generate $\textsc{down}(\textit{mouthwash})$ instead
of $\textsc{into}(\textit{mouthwash},\textit{sink})$.
Also, although a more detailed semantic representation for this sentence would
include $\textsc{put}(\textit{person},\textit{mouthwash})$, we simplify this
two-argument predicate to a one-argument predicate
$\textsc{move}(\textit{mouthwash})$, since we do not attempt to codetect people.
To ensure that we do not introduce surplus semantics, the generated predicates
always implies a \emph{weaker} constraint than the original sentence.

Each predicate is formulated around a set of primitive functions on
the arguments of the predicate.
The primitive functions produce scores indicating how well the
arguments satisfy the constraint.
The aggregate score over the functions constitutes the predicate score.
Table~\ref{tab:predicates} shows the complete list of our 24
predicates and the scores they compute.
The function $\textsf{medFlMg}(p)$ computes the median of the average
optical flow magnitude within the detections for the proposal~$p$.
The functions $\textsf{x}(p^{(t)})$ and $\textsf{y}(p^{(t)})$ return the
$x$- and $y$-coordinates of the center of $p^{(t)}$, normalized by the
frame width and height, respectively.
The function \textsf{distLessThan}$(x,a)$ is defined as
$\log[1/(1+\exp(-b(x-a)))]$, where we set $b=-20$ in the experiment.
Similarly, the function $\textsf{distGreaterThan}(x,a)$ is defined as
$\textsf{distLessThan}(-x,-a)$.
The function $\textsf{dist}(p_1^{(t)},p_2^{(t)})$ computes the
distance between the centers of $p_1^{(t)}$ and $p_2^{(t)}$, also
normalized by the frame size.
The function $\textsf{smaller}(p_1^{(t)},p_2^{(t)})$ returns 0 if the
size of $p_1^{(t)}$ is smaller than that of $p_2^{(t)}$, and $-\infty$
otherwise.
The function $\textsf{tempCoher}(p)$ evaluates whether the position of
proposal~$p$ changes during the video, by checking the position
offsets between every two frames.
A higher $\textsf{tempCoher}$ score indicates that~$p$ is more likely to be
stationary in the video.
The function $\textsf{rotAngle}(p^{(t)})$ computes the current rotated
angle of the object inside $p^{(t)}$ by looking back 1 second (30
frames).
We extract SIFT features \citep{Lowe2004} for both $p^{(t)}$ and
$p^{(t-30)}$ and match them to estimate the similarity transformation
matrix, from which the angle can be computed.
Finally the function $\textsf{hasRotation}(\alpha,\beta)$ computes the rotation
log-likelihood given angle~$\alpha$ through the von Mises distribution for
which we set the location $\mu=\beta$ and the concentration $\kappa=4$.

\begin{table}[t]
  \centering
  \resizebox{0.49\textwidth}{!}{
      \begin{tabular}{@{\hspace{0ex}}c@{\hspace{0ex}}}
        \begin{math}
          \begin{array}[t]{c@{\hspace*{30pt}}c@{\hspace*{30pt}}c}
            \Delta\textsc{distLarge}\define0.25&
            \Delta\textsc{distSmall}\define0.05&
            \Delta\textsc{angle}\define\pi/2
          \end{array}
        \end{math}\\[1.5ex]
        \hline\\[0.5ex]
        \begin{math}
          \begin{array}[t]{r@{\;\define\;}l}
            \textsc{move}(p) & \textsf{medFlMg}(p)\\
            \textsc{moveUp}(p) & \textsc{move}(p)
            + \textsf{distLessThan}\left(\textsf{y}(p^{(T)})-\textsf{y}(p^{(1)}),-\Delta\textsc{distLarge}\right)\\

            \textsc{moveDown}(p) & \textsc{move}(p)
            + \textsf{distGreaterThan}\left(\textsf{y}(p^{(T)})-\textsf{y}(p^{(1)}),\Delta\textsc{distLarge}\right)\\

            \textsc{moveVertical}(p) & \textsc{move}(p)
            + \textsf{distGreaterThan}\left(\abs{\textsf{y}(p^{(T)})-\textsf{y}(p^{(1)})},\Delta\textsc{distLarge}\right)\\

            \textsc{moveLeftwards}(p) & \textsc{move}(p)
            + \textsf{distLessThan}\left(\textsf{x}(p^{(T)})-\textsf{x}(p^{(1)}),-\Delta\textsc{distLarge}\right)\\

            \textsc{moveRightwards}(p) & \textsc{move}(p)
            + \textsf{distGreaterThan}\left(\textsf{x}(p^{(T)})-\textsf{x}(p^{(1)}),\Delta\textsc{distLarge}\right)\\

            \textsc{moveHorizontal}(p) & \textsc{move}(p)
            + \textsf{distGreaterThan}\left(\abs{\textsf{x}(p^{(T)})-\textsf{x}(p^{(1)})},\Delta\textsc{distLarge}\right)\\

            \textsc{rotate}(p) &
            \textsc{move}(p) +
            \max\limits_{t}\textsf{hasRotation}\left(\textsf{rotAngle}(p^{(t)}),\Delta\textsc{angle}\right)\\

            \textsc{towards}(p_1,p_2) & \textsc{move}(p_1)
            +\textsf{distLessThan}\left(\textsf{dist}(p_1^{(T)},p_2^{(T)})
            -\textsf{dist}(p_1^{(1)},p_2^{(1)}),-\Delta\textsc{distLarge}\right)\\

            \textsc{awayFrom}(p_1,p_2) & \textsc{move}(p_1)
            +\textsf{distGreaterThan}\left(\textsf{dist}(p_1^{(T)},p_2^{(T)})
            -\textsf{dist}(p_1^{(1)},p_2^{(1)}),\Delta\textsc{distLarge}\right)\\

            \textsc{leftOfStart}(p_1,p_2) & \textsf{tempCoher}(p_2)
            +\textsf{distLessThan}\left(\textsf{x}(p_1^{(1)})-\textsf{x}(p_2^{(1)}),
            -\Delta\textsc{distSmall}\right)\\

            \textsc{leftOfEnd}(p_1,p_2) & \textsf{tempCoher}(p_2)
            +\textsf{distLessThan}\left(\textsf{x}(p_1^{(T)})-\textsf{x}(p_2^{(T)}),
            -\Delta\textsc{distSmall}\right)\\

            \textsc{rightOfStart}(p_1,p_2) & \textsf{tempCoher}(p_2)
            +\textsf{distGreaterThan}\left(\textsf{x}(p_1^{(1)})-\textsf{x}(p_2^{(1)}),
            \Delta\textsc{distSmall}\right)\\

            \textsc{rightOfEnd}(p_1,p_2) & \textsf{tempCoher}(p_2)
            +\textsf{distGreaterThan}\left(\textsf{x}(p_1^{(T)})-\textsf{x}(p_2^{(T)}),
            \Delta\textsc{distSmall}\right)\\

            \textsc{onTopOfStart}(p_1,p_2) &
            \begin{array}[t]{@{}l@{}}
              \textsf{tempCoher}(p_2)\\
              +\textsf{distGreaterThan}\left(\textsf{y}(p_1^{(1)})-\textsf{y}(p_2^{(1)}),
              -2\Delta\textsc{distLarge}\right)\\
              +\textsf{distLessThan}\left(\textsf{y}(p_1^{(1)})-\textsf{y}(p_2^{(1)}),0\right)\\
              +\textsf{distLessThan}\left(\abs{\textsf{x}(p_1^{(1)})-\textsf{x}(p_2^{(1)})},2\Delta\textsc{distSmall}\right)\\
            \end{array}\\
            \textsc{onTopOfEnd}(p_1,p_2) &
            \begin{array}[t]{@{}l@{}}
              \textsf{tempCoher}(p_2)\\
              +\textsf{distGreaterThan}\left(\textsf{y}(p_1^{(T)})-\textsf{y}(p_2^{(T)}),
              -2\Delta\textsc{distLarge}\right)\\
              +\textsf{distLessThan}\left(\textsf{y}(p_1^{(T)})-\textsf{y}(p_2^{(T)}),0\right)\\
              +\textsf{distLessThan}\left(\abs{\textsf{x}(p_1^{(T)})-\textsf{x}(p_2^{(T)})},2\Delta\textsc{distSmall}\right)\\
            \end{array}\\
            \textsc{nearStart}(p_1,p_2) & \textsf{tempCoher}(p_2)
            +\textsf{distLessThan}\left(\textsf{dist}(p_1^{(1)},p_2^{(1)}),2\Delta\textsc{distSmall}\right)\\

            \textsc{nearEnd}(p_1,p_2) & \textsf{tempCoher}(p_2)
            +\textsf{distLessThan}\left(\textsf{dist}(p_1^{(T)},p_2^{(T)}),2\Delta\textsc{distSmall}\right)\\

            \textsc{inStart}(p_1,p_2) & \textsf{tempCoher}(p_2)
            + \textsc{nearStart}(p_1,p_2) + \textsf{smaller}(p_1^{(1)},p_2^{(1)})\\

            \textsc{inEnd}(p_1,p_2) & \textsf{tempCoher}(p_2)
            + \textsc{nearEnd}(p_1,p_2) + \textsf{smaller}(p_1^{(T)},p_2^{(T)})\\

            \textsc{belowStart}(p_1,p_2) & \textsf{tempCoher}(p_2)
            + \textsf{distGreaterThan}\left(\textsf{y}(p_1^{(1)})-\textsf{y}(p_2^{(1)}),\Delta\textsc{distSmall}\right)\\

            \textsc{belowEnd}(p_1,p_2) & \textsf{tempCoher}(p_2)
            + \textsf{distGreaterThan}\left(\textsf{y}(p_1^{(T)})-\textsf{y}(p_2^{(T)}),\Delta\textsc{distSmall}\right)\\

            \textsc{aboveStart}(p_1,p_2) & \textsf{tempCoher}(p_2)
            + \textsf{distLessThan}\left(\textsf{y}(p_1^{(1)})-\textsf{y}(p_2^{(1)}),-\Delta\textsc{distSmall}\right)\\

            \textsc{aboveEnd}(p_1,p_2) & \textsf{tempCoher}(p_2)
            + \textsf{distLessThan}\left(\textsf{y}(p_1^{(T)})-\textsf{y}(p_2^{(T)}),-\Delta\textsc{distSmall}\right)\\

            \textsc{over}(p_1,p_2) &
            \begin{array}[t]{@{}l@{}}
              \textsf{tempCoher}(p_2)\\
              + \max\limits_{t}
              \left(\begin{array}{@{}l@{}}
                \textsf{distLessThan}\left(\textsf{y}(p_1^{(t)})-\textsf{y}(p_2^{(t)}),-\Delta\textsc{distSmall}\right)\\
                \textsf{distLessThan}\left(\abs{\textsf{x}(p_1^{(t)})-\textsf{x}(p_2^{(t)})},\Delta\textsc{distLarge}\right)\\
              \end{array}\right)\\
            \end{array}\\
          \end{array}
        \end{math}\\
      \end{tabular}
  }
  \vspace*{1ex}
  \caption{Our predicates and their semantics.
    For simplicity, we show the computation on only a single first frame
    $p^{(1)}$ or last frame $p^{(T)}$ of a proposal.
    In practice, to reduce noise, all of the scores are averaged over the
    first or last~$L$ frames.}
  \label{tab:predicates}
\end{table}

\subsection{Generating Object Proposals}

We first generate~$N$ object candidates for each video frame.
We use EdgeBoxes \citep{Zitnick2014} to obtain the $\frac{N}{2}$ top-ranking
object candidates and MCG \citep{Arbelaez2014} to obtain the other
half, filtering out candidates larger than $\frac{1}{20}$ of the
video-frame size to focus on small and medium-sized objects.
This yields~$NT$ object candidates for a video with~$T$ frames.
We then generate~$K$ object proposals from these~$NT$ candidates.
To obtain object proposals with object candidates of consistent appearance and
spatial location, one would nominally require that $K \ll N^T$.
To circumvent this, we first randomly sample a frame~$t$ from the video with
probability proportional to the averaged magnitude of optical flow
\citep{Farneback2003} within that frame.
Then, we sample an object candidate from the~$N$ candidates in frame~$t$.
To decide whether the object is moving or not, we sample from
\{\textsc{moving},\textsc{stationary}\} with distribution
$\{\frac{1}{3},\frac{2}{3}\}$.
We sample a \textsc{moving} object candidate with probability
proportional to the average flow magnitude within the candidate.
Similarly, we sample a \textsc{stationary} object candidate with probability
inversely proportional to the average flow magnitude within the
candidate.
The sampled candidate is then propagated (tracked) bidirectionally to
the start and the end of the video.
We use the CamShift algorithm \citep{Bradski1998} to track both
\textsc{moving} and \textsc{stationary} objects, allowing the size of
\textsc{moving} objects to change during the process, but requiring the size of
\textsc{stationary} objects to remain constant.
\textsc{Stationary} objects are tracked to account for noise or occlusion that
manifests as small motion or change in size.
We track \textsc{moving} objects in HSV color space and \textsc{stationary}
objects in RGB color space.
We do not use optical-flow-based tracking methods since these methods suffer
from drift when objects move quickly.
We repeat this sampling and propagation process~$K$ times to obtain~$K$ object
proposals $\{p_k\}$ for each video.
Examples of the sampled proposals ($K=240$) are shown in the middle column of
Figure~\ref{fig:pipeline}.

\subsection{Similarity between Object Proposals}
\label{sec:similarity}

We compute the appearance similarity of two object proposals as follows.
We first uniformly sample~$M$ detections $\{b^m\}$ from each proposal
along its temporal extent.
For each sampled detection, we extract PHOW \citep{Bosch2007} and HOG
\citep{Dalal2005} features to represent its appearance and shape.
We also do so after we rotate this detection by 90\degr, 180\degr, and 270\degr.
Then, we measure the similarity~$g$ between a pair of detections~$b_1^m$
and~$b_2^m$ with:
\begin{equation*}
\begin{array}{r}
  g(b_1^m,b_2^m)=\max\limits_{i,j\in\{0,1,2,3\}}\frac{1}{2}\left(\begin{array}{l}
  g_{\chi^2}(\rot_i(b_{1}^m),\rot_j(b_{2}^m))\\
  {}+g_{L_2}(\rot_i(b_{1}^m),\rot_j(b_{2}^m))\end{array}\right)\\
\end{array}
\end{equation*}
where $\rot_i\;i=0,1,2,3$ represents rotation by 0\degr, 90\degr, 180\degr, and
270\degr, respectively.
We use $g_{\chi^2}$ to compute the $\chi^2$ distance between the PHOW
features and $g_{L_2}$ to compute the Euclidean distance between
the HOG features, after which the distances are linearly scaled to $[0,1]$
and converted to log similarity scores.
Finally, the similarity between two proposals~$p_1$ and~$p_2$ is taken to be:
\[g(p_1,p_2)=\mathop{\text{median}}\limits_{m}g(b_1^m,b_2^m)\]

\begin{table*}[t]
  \centering
  \resizebox{\textwidth}{!}{
    \begin{tabular}{@{}l||l|l|l|l|l|l|l|l|l|l||l|l|l|l|l||@{}}
      & \multicolumn{10}{c||}{Our new dataset, codetection set \#} &
      \multicolumn{5}{c||}{Our subset of CAD-120, codetection set \#}\\
      \cline{2-16}
      & 1 & 2 & 3 & 4 & 5 & 6 & 7 & 8 & 9 & 10 & 1 & 2 & 3 & 4 & 5\\
      \hline
      Scene & \textsc{k1} &
      \textsc{k2} & \textsc{k2,3} & \textsc{k4} & \textsc{b}
      & \textsc{b} & \textsc{g} & \textsc{k1,2,3} & \textsc{b\&g}
      & \textsc{k1,2,3,4} & \multicolumn{5}{c||}{}\\
      \hline
      \hline
      \multirow{8}{*}{Objects} & \emph{box} & \emph{bowl} & \emph{bowl}
      & \emph{cup} & \emph{box} & \emph{box} & \emph{bucket} &
      \emph{bowl} & \emph{box} & \emph{bowl} & \emph{bowl} & \emph{bowl} &
      \emph{bowl} & \emph{bowl} & \emph{bowl}\\ 
      & \emph{cabbage} & \emph{cabbage} & \emph{cabbage} & \emph{juice}
      & \emph{cooler} & \emph{cooler} & \emph{gas can}
      & \emph{cabbage} & \emph{bucket} & \emph{cabbage} & \emph{cereal} &
      \emph{cereal} & \emph{cereal} & \emph{cereal} & \emph{cereal}\\ 
      & \emph{coffee grinder} & \emph{coffee grinder} & \emph{pineapple}
      & \emph{ketchup} & & & \emph{watering pot} & \emph{pineapple}
      & \emph{cooler} & \emph{cup} & \emph{cup} & \emph{cup} & \emph{cup} &
      \emph{cup}& \emph{cup}\\ 
      & \emph{mouthwash} & \emph{mouthwash} & \emph{squash} &
      \emph{milk}
      & & & & \emph{squash} & \emph{gas can} & \emph{juice} & \emph{jug} &
      \emph{jug} & \emph{jug} & \emph{jug} & \emph{jug}\\ 
      & \emph{pineapple} & \emph{pineapple} & & & & & &
      & \emph{watering pot} & \emph{ketchup} & \emph{microwave} &
      \emph{microwave} & \emph{microwave} & \emph{microwave} & \emph{microwave}\\ 
      & \emph{squash} & \emph{squash} & & & & & & & & \emph{milk} & & & & &
      \\ 
      & & & & & & & & & & \emph{mouthwash} & & & & &\\ 
      & & & & & & & & & & \emph{pineapple} & & & & &\\ 
      \hline
      \# of videos & 26 & 27 & 17 & 21 & 19 & 17 & 23 & 17 & 25 & 24 & 15 & 15
      & 15 & 15 & 15\\
      \hline
      \# vertices in run 1 & 33 & 29 & 24 & 41 & 25 & 26 & 32 & 21 & 35 & 39 &
      29 & 27 & 24 & 26 & 27\\
      \# vertices in run 2 & 34 & 37 & 32 & 46 & 24 & 22 & 27 & 26 & 32 & 41 &
      25 & 27 & 24 & 22 & 27\\
      \# vertices in run 3 & 33 & 38 & 31 & 36 & 24 & 22 & 33 & 27 & 35 & 39 &
      25 & 26 & 21 & 23 & 26\\
      \hline
    \end{tabular}}
  \vspace*{1ex}
  \caption{The experimental setup of the 10 codetection sets for our new dataset
    and the 5 codetection sets for our subset of CAD-120.}
  \label{tab:sets}
\end{table*}

\subsection{Joint Inference}

We extract object instances (see all 15 classes for our new dataset and all 5
classes for our subset of CAD-120 in Section~\ref{sec:experiment}) from the
sentences and model them as vertices in a graph.
Each vertex~$v$ can be assigned one of the~$K$ proposals in the video
that is paired with the sentence in which the vertex occurs.
The score of assigning a proposal~$k_v$ to a vertex~$v$ is taken to be
the unary predicate score~$h_v(k_v)$ computed from the sentence (if
such exists, or otherwise 0).
We construct an edge between every two vertices~$v$ and~$u$ that belong to the
same object class.
We denote this class membership relation as $(v,u)\in\mathcal{C}$.
The score of this edge $(v,u)$, when the proposal~$k_v$ is assigned to
vertex~$v$ and the proposal~$k_u$ is assigned to vertex~$u$, is taken to be the
similarity score $g_{v,u}(k_v,k_u)$ between the two proposals, as described in
Section~\ref{sec:similarity}.
Similarly, we also construct an edge between two vertices~$v$ and~$u$ that are
arguments of the same predicate.
We denote this predicate membership relation as $(v,u)\in\mathcal{P}$.
The score of this edge $(v,u)$, when the proposal~$k_v$ is assigned to
vertex~$v$ and the proposal~$k_u$ is assigned to vertex~$u$, is taken to be the
predicate score $h_{v,u}(k_v,k_u)$ between the two proposals, as described in
Section~\ref{sec:predicate}.
Our problem, then, is to select a proposal for each vertex that
maximizes the joint score on this graph, \ie\ solving the following
optimization problem:
\begin{equation*}
  \max\limits_{\mathbf{k}}\sum_vh_v(k_v)+\sum_{(v,u)\in\mathcal{C}}g_{v,u}(k_v,k_u)+\sum_{(v,u)\in\mathcal{P}}h_{v,u}(k_v,k_u)
\end{equation*}
where~$\mathbf{k}$ is the collection of the selected proposals for all
the vertices.
Note that the unary and binary scores are equally weighted in the above
objective function.
This discrete inference problem on graphical models can be solved approximately
by Belief Propagation \citep{Pearl1982}.
In the experiment, we use the OpenGM \citep{Andres2012} implementation
to find the approximate solution.

\section{Experiment}
\label{sec:experiment}

\begin{figure*}[t]
  \setlength{\tabcolsep}{3pt}
  \centering
  \resizebox{\textwidth}{!}{
    \begin{tabular}{@{}ccccccccccccccc@{\hspace*{10pt}}|@{\hspace*{10pt}}ccccc@{}}
      \includegraphics[width=0.12\textwidth]{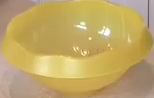}&
      \includegraphics[width=0.12\textwidth]{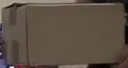}&
      \includegraphics[width=0.12\textwidth]{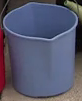}&
      \includegraphics[width=0.12\textwidth]{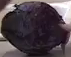}&
      \includegraphics[width=0.12\textwidth]{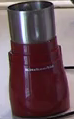}&
      \includegraphics[width=0.12\textwidth]{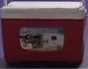}&
      \includegraphics[width=0.12\textwidth]{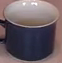}&
      \includegraphics[width=0.12\textwidth]{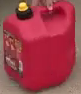}&
      \includegraphics[width=0.12\textwidth]{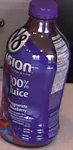}&
      \includegraphics[width=0.12\textwidth]{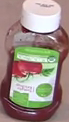}&
      \includegraphics[width=0.12\textwidth]{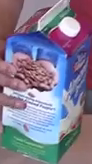}&
      \includegraphics[width=0.12\textwidth]{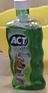}&
      \includegraphics[width=0.12\textwidth]{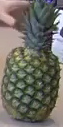}&
      \includegraphics[width=0.12\textwidth]{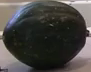}&
      \includegraphics[width=0.12\textwidth]{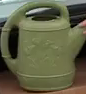}&
      \includegraphics[width=0.12\textwidth]{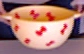}&
      \includegraphics[width=0.12\textwidth]{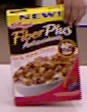}&
      \includegraphics[width=0.12\textwidth]{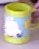}&
      \includegraphics[width=0.12\textwidth]{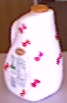}&
      \includegraphics[width=0.12\textwidth]{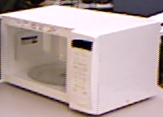}\\
    \end{tabular}}
  \caption{Examples of the 15 object classes to be codetected in our new
    dataset and the 5 object classes to be codetected in our subset of CAD-120.
    From left to right: the object classes in our new dataset, \emph{bowl},
    \emph{box}, \emph{bucket}, \emph{cabbage}, \emph{coffee grinder},
    \emph{cooler}, \emph{cup}, \emph{gas can}, \emph{juice}, \emph{ketchup},
    \emph{milk}, \emph{mouthwash}, \emph{pineapple}, \emph{squash}, and
    \emph{watering pot}, and the object classes in our subset of CAD-120,
    \emph{bowl}, \emph{cereal}, \emph{cup}, \emph{jug}, and \emph{microwave}.}
  \label{fig:20objects}
\end{figure*}

Our method can only be applied to datasets with the following properties:
\begin{compactenum}[I)]
\item It can only be applied to video that depicts motion and changing spatial
  relations between objects.
  \label{item:I}
\item It can only be applied to video, not images, because it relies on such
  motion and changing spatial relations.
  \label{item:II}
\item The video must be paired with sentences that describe that motion and
  those changing spatial relations.
  Some existing image and video corpora are paired with sentences that do not
  describe such.
  \label{item:III}
\item The objects to be codetected must be detectable by existing object
  proposal methods.
  \label{item:IV}
\item There must be different clips that all involve different instances of
  the same object class participating in the described activity.
  This is necessary to support \emph{codetection}.
  \label{item:V}
\end{compactenum}
Most existing datasets do not meet the above criteria and are thus ill suited
to our approach.
We evaluate on two specific datasets that are suited.
Most existing methods require properties that our datasets lack.
For example, \citet{Srikantha2014} require depth and human pose information.
Others, such as \citet{Prest2012}, \citet{Schulter2013}, \citet{Armand2014},
and \citet{Ramanathan2014} do not make code available.
Thus neither can one run our method on existing datasets or existing methods
on our datasets.

It is not possible to compare our method to existing image codetection methods
or evaluate on existing image codetection datasets, or any existing image
captioning datasets, because they lack properties~\ref{item:I}, \ref{item:II},
and~\ref{item:III}.
Further, it is not possible to compare our method to existing video codetection
methods or existing video codetection datasets.
\citet{Schulter2013} and \citet{Ramanathan2014} address different problems with
datasets that are highly specific to those problems and are thus incomparable.
The dataset used by \citet{Prest2012} and \citet{Armand2014} lacks
properties~\ref{item:I} and~\ref{item:III}.
\citet{Srikantha2014} evaluate on three datasets: ETHZ-activity
\citep{Fossati2013}, CAD-120, and MPII-cooking \citep{Rohrbach2012}.
Two of the these, namely ETHZ-activity and MPII-cooking, lack
properties~\ref{item:III} and~\ref{item:IV}.
\citet{Srikantha2014} rely on depth and human pose information to overcome
the lack of property~\ref{item:IV}.
Moreover, the kinds of activity depicted in ETHZ-activity and MPII-cooking
cannot easily be formulated in terms of descriptions of object motion and
changing spatial relations.
We do apply our method to a subset of CAD-120.
However, because we do not use depth and human pose information, we only
consider that subset of CAD-120 that satisfies property~\ref{item:IV}.
\citet{Srikantha2014} apply their method to a different subject, rendering
their results incomparable with ours.
Moreover, we use incompatible sources of information: we use sentences but they
do not; they use depth and human pose but we do not.
This it is impossible to perform an apples-to-apples comparison, even on the
common subset.

There exist a large number of video datasets that are not used for codetection
but rather are used for others purposes like activity recognition and video
captioning.
Sentential annotation is available for some of these, like MPII-cooking,
M-VAD \citep{Torabi2015}, and MPII-MD \citep{Rohrbach2015}.
However, the vast majority of the clips in M-VAD (48,986 clips annotated
with sentences) and MPII-MD (68,337 clips annotated with sentences) do not
satisfy properties~\ref{item:I} and~\ref{item:IV}.
We searched the sentential annotations from each of these two corpora for all
instances of twelve common English verbs that represent the kinds of verbs whose
that describe motion and changing spatial relations between object.
\begin{center}
  \begin{tabular}{l|rr|rr}
    & \multicolumn{2}{r}{M-VAD} & \multicolumn{2}{r}{MPII-MD}\\
    \hline
    \emph{add}    &   89 & 0/10 &  120 & 0/10\\
    \emph{carry}  &   74 & 1/10 &  273 & 2/10\\
    \emph{lift}   &  435 & 1/10 &  374 & 0/10\\
    \emph{load}   &   48 & 0/10 &   89 & 0/10\\
    \emph{move}   &  332 & 0/10 & 1106 & 0/10\\
    \emph{pick}   &  366 & 1/10 &  703 & 1/10\\
    \emph{pour}   &   95 & 0/10 &  207 & 1/10\\
    \emph{put}    &  294 & 1/10 &  921 & 0/10\\
    \emph{rotate} &   27 & 0/10 &   13 & 0/10\\
    \emph{stack}  &   91 & 0/10 &   56 & 0/10\\
    \emph{take}   & 1058 & 0/10 & 1786 & 0/10\\
    \emph{unload} &    1 & 0/10 &   11 & 2/10
  \end{tabular}
\end{center}
We further examined ten sentences for each verb from each corpus, together with
the corresponding video clips, and found that only ten out of the 240 examined
satisfied properties~\ref{item:I} and~\ref{item:IV}.
Moreover, none of these ten suitable video clips satisfied
property~\ref{item:V}.
Further, of the twelve classes (\emph{AnswerPhone}, \emph{DriveCar},
\emph{Eat}, \emph{FightPerson}, \emph{GetOutCar}, \emph{HandShake},
\emph{HugPerson}, \emph{Kiss}, \emph{Run}, \emph{SitDown}, \emph{SitUp}, and
\emph{StandUp}) in the Hollywood~2 dataset \citep{Marszalek2009}, only four
(\emph{AnswerPhone}, \emph{DriveCar}, \emph{GetOutCar}, and \emph{Eat}) satisfy
property~\ref{item:I}.
Of these, three classes (\emph{AnswerPhone}, \emph{DriveCar}, and
\emph{GetOutCar}) always depict a single object class, and thus are ill suited
for codetecting anything but the two fixed classes \emph{phone} and \emph{car}.
The one remaining class (\emph{Eat}) fails to satisfy property~\ref{item:V}.
This same situation occurs with essentially all standard datasets used for
activity recognition, like UCF Sports \citep{Rodriguez2008}.

The standard sources of naturally occurring video for corpora used within the
computer-vision community are Hollywood movies and YouTube video clips.
However, Hollywood movies, in general, mostly involve dialog among actors, or
generic scenery and backgrounds.
At best, only small portions of most Hollywood movies satisfy
property~\ref{item:I}, and such rarely is reflected in the dialog or script,
thus failing to satisfy property~\ref{item:III}.
We attempted to gather a codetection corpus from YouTube.
But again, about a dozen students searching YouTube for depictions of about a
dozen common English verbs, examining hundreds of hits, found that fewer than
1\% satisfied property~\ref{item:I} and non satisfied property~\ref{item:V}.
Thus it is only feasible to evaluate our method on video that has been filmed
to expressly satisfy properties~\ref{item:I}--\ref{item:V}.

While existing datasets within the computer-vision community do not
satisfy properties~\ref{item:I}--\ref{item:V}, we believe that these properties
are nonetheless reflective of the real natural world.
In the real world, people interact with everyday objects (in their kitchen,
basement, driveway, and many similar location) all of the time.
It is just that people don't usually record such video, let alone make
Hollywood movies about it or post it on YouTube.
Further, people rarely describe such in naturally occurring text in movie
scripts or in text uploaded to YouTube.
Yet, children---and even adults---probably learn names of newly observed objects
by observing people in their environment interacting with those objects in the
context of dialog about such.
Thus we believe that our problem, and our datasets, are a natural reflection of
the kinds of learning that people employ to learn to recognize newly named
objects.

\subsection{Datasets}

We evaluate our method on two datasets that do satisfy these properties.
The first is a newly collected dataset, filmed to expressly satisfy
properties~\ref{item:I}--\ref{item:V}.
This dataset was filmed in 6 different scenes (four in the
\textsc{kitchen}, one in the \textsc{basement}, and one outside the
\textsc{garage}) of a house.
The lighting conditions vary greatly across the different scenes, with the
\textsc{basement} scene the darkest, the \textsc{kitchen} scene exhibiting
modest lighting, and the \textsc{garage} scene the brightest.
Within each scene, the lighting often varies across different video regions.
We assigned 5 actors (four adults and one child) with 15 distinct everyday
objects (\emph{bowl}, \emph{box}, \emph{bucket}, \emph{cabbage}, \emph{coffee
  grinder}, \emph{cooler}, \emph{cup}, \emph{gas can}, \emph{juice},
\emph{ketchup}, \emph{milk}, \emph{mouthwash}, \emph{pineapple}, \emph{squash},
and \emph{watering pot}, see Figure~\ref{fig:20objects}), and had them perform
different actions which involve interaction with these objects.
No special instructions were given requiring that the actors move slowly or the
that objects not be occluded.
The actors often are partially outside the field of view.
Note that the dataset used by \citet{Srikantha2014} does not exhibit this
property.
Indeed, their method employs human pose which requires that the human be
sufficiently visible to estimate such.
The filming was performed using a normal consumer camera that introduces motion
blur on the objects when the actors move quickly.
We downsampled the filmed videos to $\text{768}\times\text{432}$ and divided
them into 150 short video clips, each clip depicting a specific event lasting
between~2 and~6 seconds at 30~fps.
The 150 video clips constitute a total of 12,509 frames.

The second dataset is a subset of of CAD-120.
Many of the 120 clips in CAD-120 depict sequences of actions.
We divide those clips into subclips, each containing one action.
We discard those that fail to satisfy any of the properties~\ref{item:I},
\ref{item:IV}, or~\ref{item:V}, leaving 75 clips.
These clips have spatial resolution $\text{640}\times\text{480}$, each clip
depicting a specific event lasting between~3 and~5 seconds at 30~fps.
The 75 video clips constitute a total of 8,854 frames, and contain 5 distinct
object classes, namely \emph{bowl}, \emph{cereal}, \emph{cup}, \emph{jug}, and
\emph{microwave}.

\subsection{Experimental Setup}

We employed Amazon Mechanical Turk
(AMT)\footnote{\texttt{\url{https://www.mturk.com/mturk/}}} to obtain three
distinct sentences, by three different workers, for each video clip in each
dataset, resulting in 450 sentences for the our new dataset and 225 sentences
for our subset of CAD-120.
AMT annotators were simply instructed to provide a single sentence for each
video clip that described the primary activity depicted taking place with
objects from a common list of object classes that occur in the entire dataset.
The collected sentences were all converted to the predicates in
Table~\ref{tab:predicates} using the methods of Section~\ref{sec:predicate}.
We processed each of the two datasets three times, each time using a different
set of sentences produced by different workers; each sentence was used in
exactly one run of the experiment.
Furthermore, we divided each corpus into codetection set, each set
containing a small subset of the video-sentence pairs.
(For our new dataset, some pairs were reused in different codetection sets.
For CAD-120, each pair was used in exactly one codetection set.)
Some codetection sets contained only videos filmed in the same background,
while others contained a mix of videos filmed in different backgrounds.
(The backgrounds in each codetection set for our new dataset are summarized in
Table~\ref{tab:sets}, where \textsc{k}, \textsc{b}, and \textsc{g} denote
\textsc{kitchen}, \textsc{basement}, and \textsc{garage}, respectively.)
This rules out the possibility of codetecting objects by simple
background modeling (\eg\ background subtraction).
Codetection sets were processed independently, each with a distinct graphical
model.
Table~\ref{tab:sets} contains the number of video-sentence pairs and the
number of vertices in the resulting graphical model for each codetection set of
each corpus.

We compared the resulting codetections against human annotation.
Human-annotated boxes around objects are provided with CAD-120.
For our new dataset, these were obtained with AMT.\@
We obtained five bounding-box annotations for each target object in each video
frame.
We asked annotators to annotate the referent of a specific highlighted word in
the sentence associated with the video containing that frame.
Thus the annotation reflects the semantic constraint implied by the sentences.
This resulted in $\text{5}\times\text{289}=\text{1445}$ human annotated tracks.
To measure how well codetections match human annotation, we use the
\emph{IoU}, namely the ratio of the area of the intersection of two bounding
boxes to the area of their union.
The object codetection problem exhibits inherent ambiguity: different
annotators tend to annotate different parts of an object or make different
decisions whether to include surrounding background regions when the object is
partially occluded.
To quantify such ambiguity, we computed intercoder agreement between the human
annotators for our datasets.
We computed $\frac{\text{5}\times{4}}{2}=\text{10}$ IoU scores for all box
pairs produced by the 5 annotators in every frame and averaged them over the
entire dataset, obtaining an overall human-human IoU of 0.72.\footnote{Both
  datasets, including videos, sentences, and bounding-box annotations, are
  available at
  \texttt{\url{http://upplysingaoflun.ecn.purdue.edu/~qobi/cccp/sentence-codetection.html}}}.

We found no publicly available implementations of existing video object
codetection methods \citep{Prest2012, Schulter2013, Armand2014, Srikantha2014,
  Ramanathan2014},
thus for comparison we employ four variants of our method that alternatively
disable different scores in our codetection framework.
These variants help one understand the relative importance of different
components of the framework.
Together with our full method, they are summarized below:

\begin{center}
\resizebox{0.48\textwidth}{!}{
  \begin{tabular}{l|l|l|l|l|l}
    \hline
    & \textbf{SIM} & \textbf{FLOW} & \textbf{SENT} & \textbf{SIM+FLOW}
    & \textbf{SIM+SENT}\\
    & & & & & (our full method)\\
    \hline
    \hline
    Flow score? & no & yes & yes & yes & yes\\
    Similarity score? & yes & no & no & yes & yes\\
    Sentence score? & no & partial & yes & partial & yes\\
    \hline
  \end{tabular}}
\end{center}

\begin{figure*}[t]
  \centering
  \resizebox{\textwidth}{!}{
    \begin{tabular}{ccc}
      \includegraphics[width=0.33\textwidth]{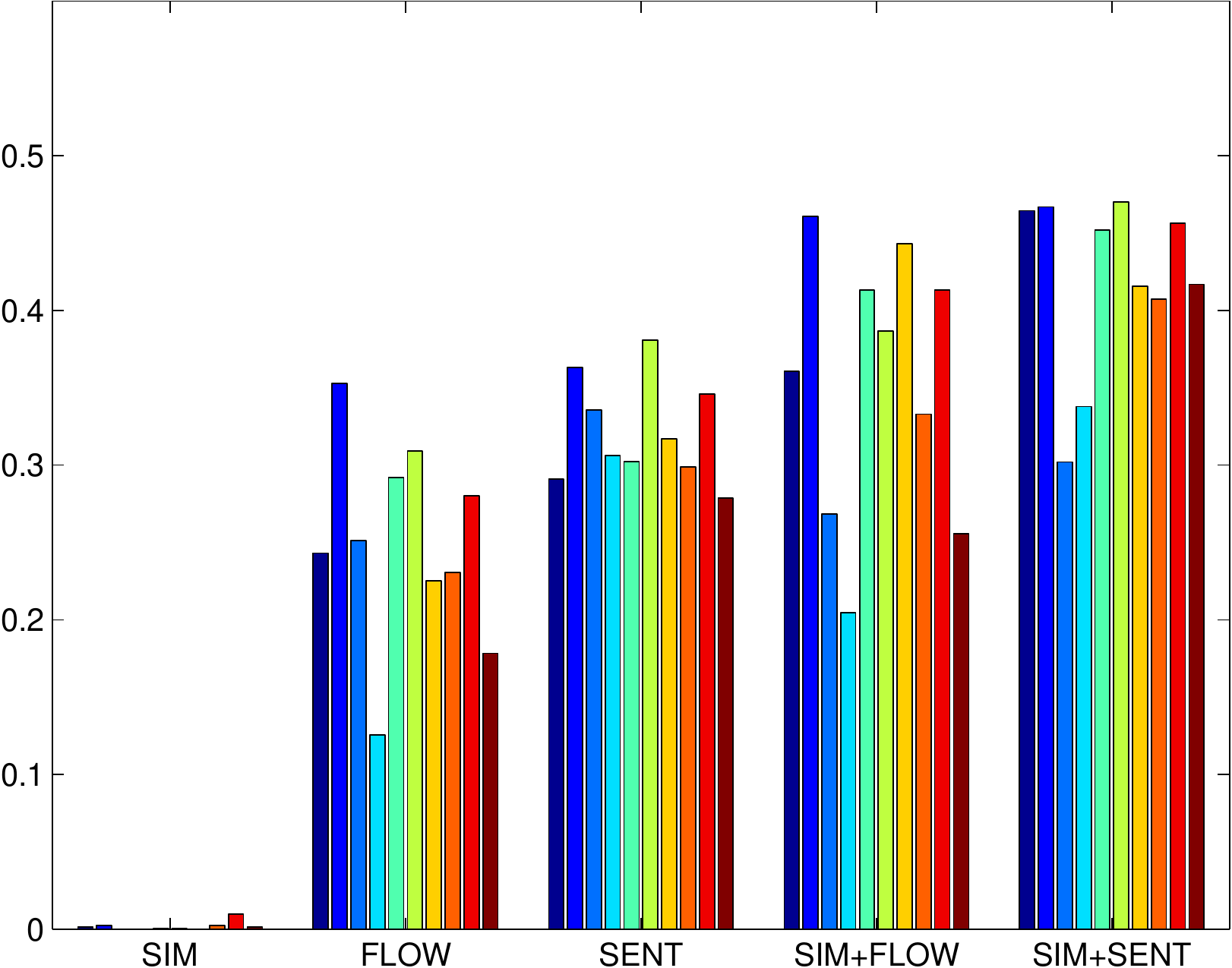}&
      \includegraphics[width=0.33\textwidth]{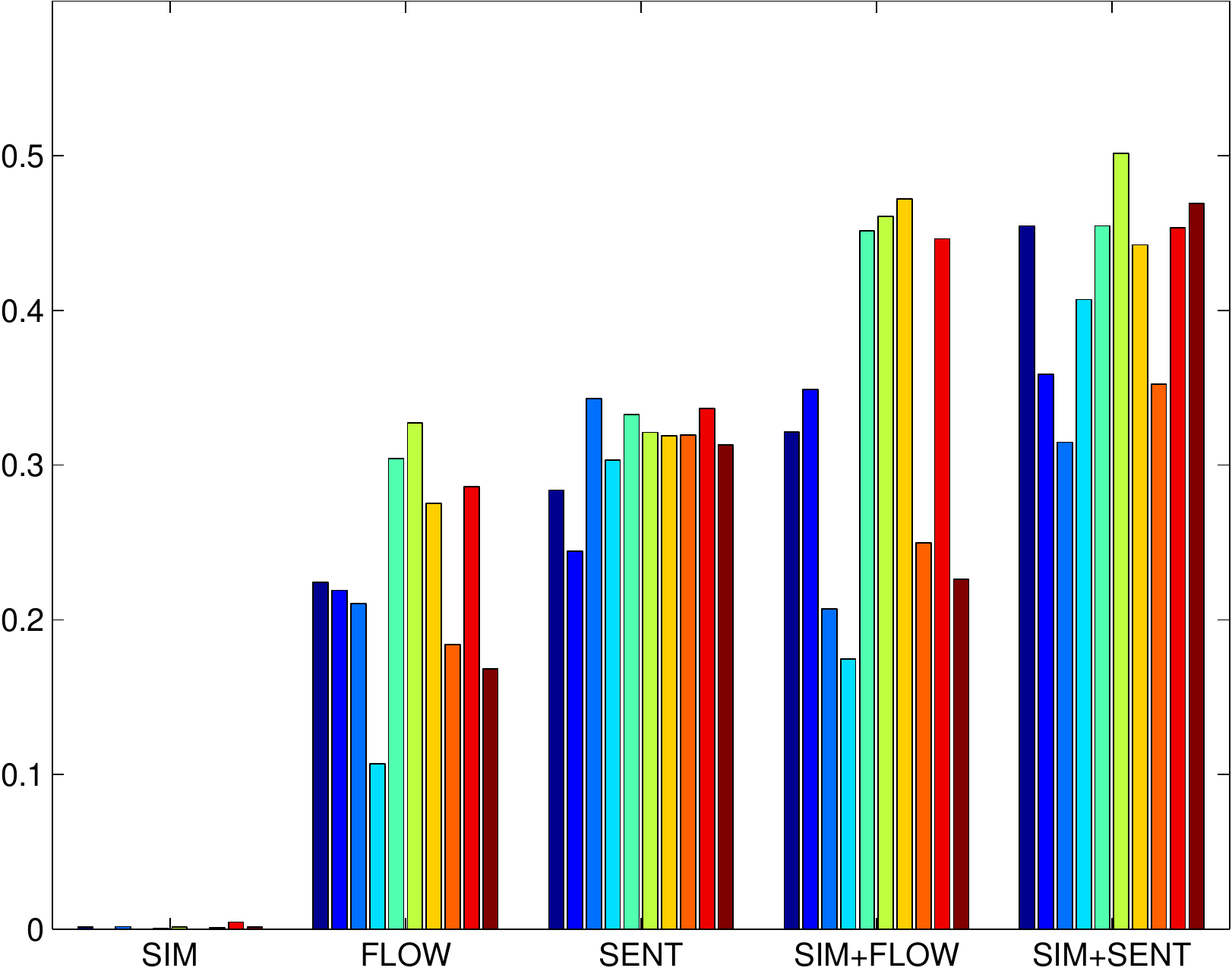}&
      \includegraphics[width=0.33\textwidth]{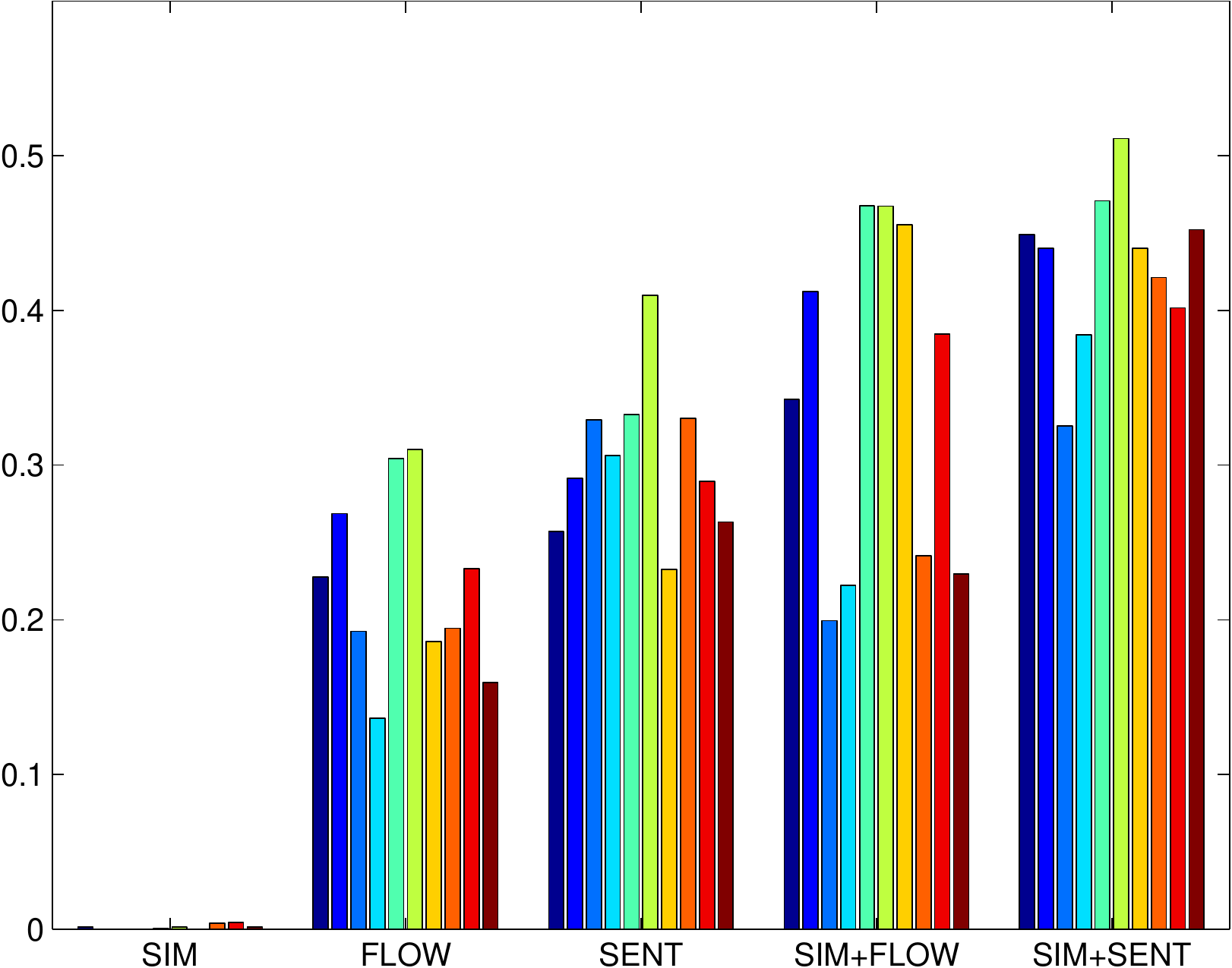}\\
      \includegraphics[width=0.33\textwidth]{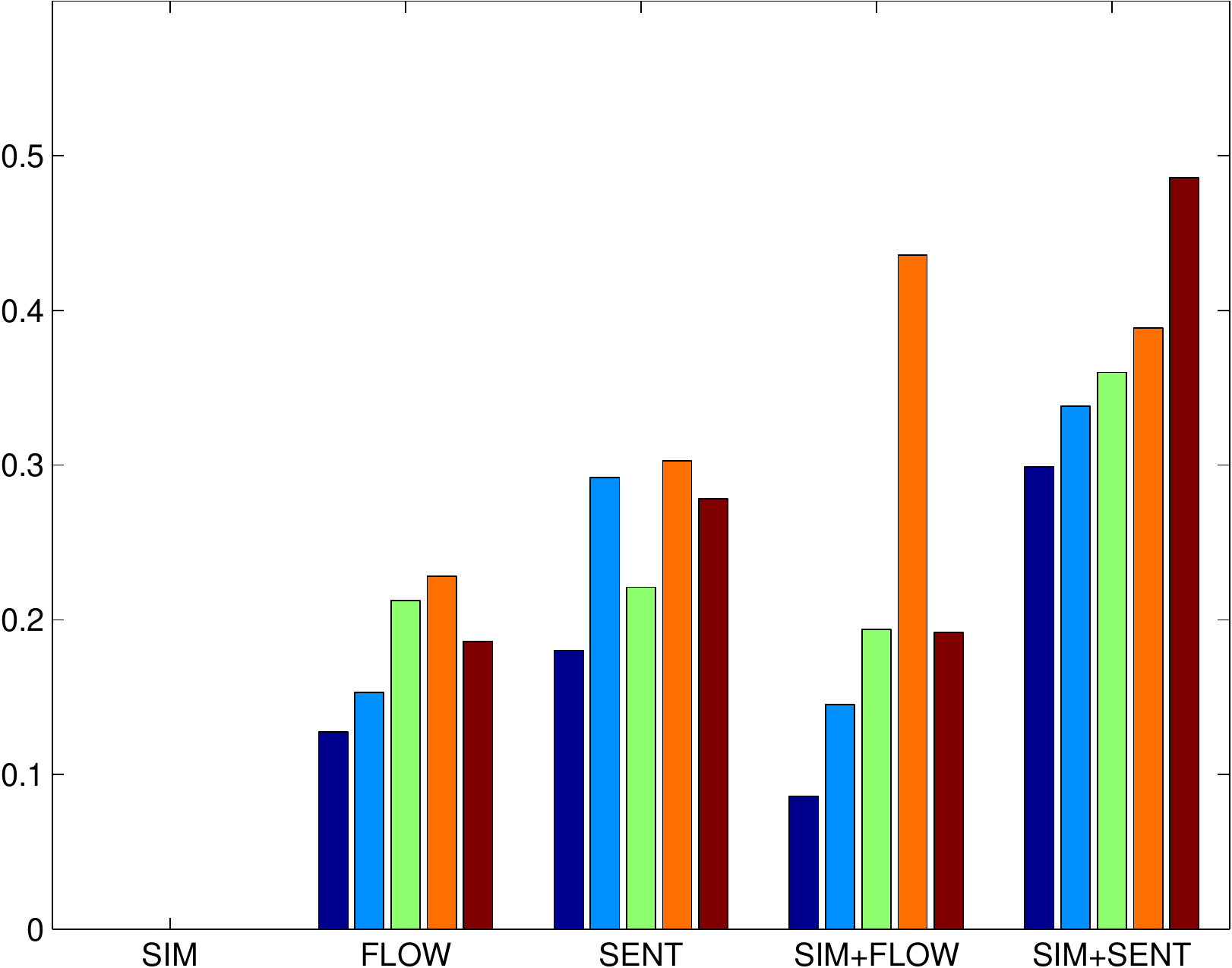}&
      \includegraphics[width=0.33\textwidth]{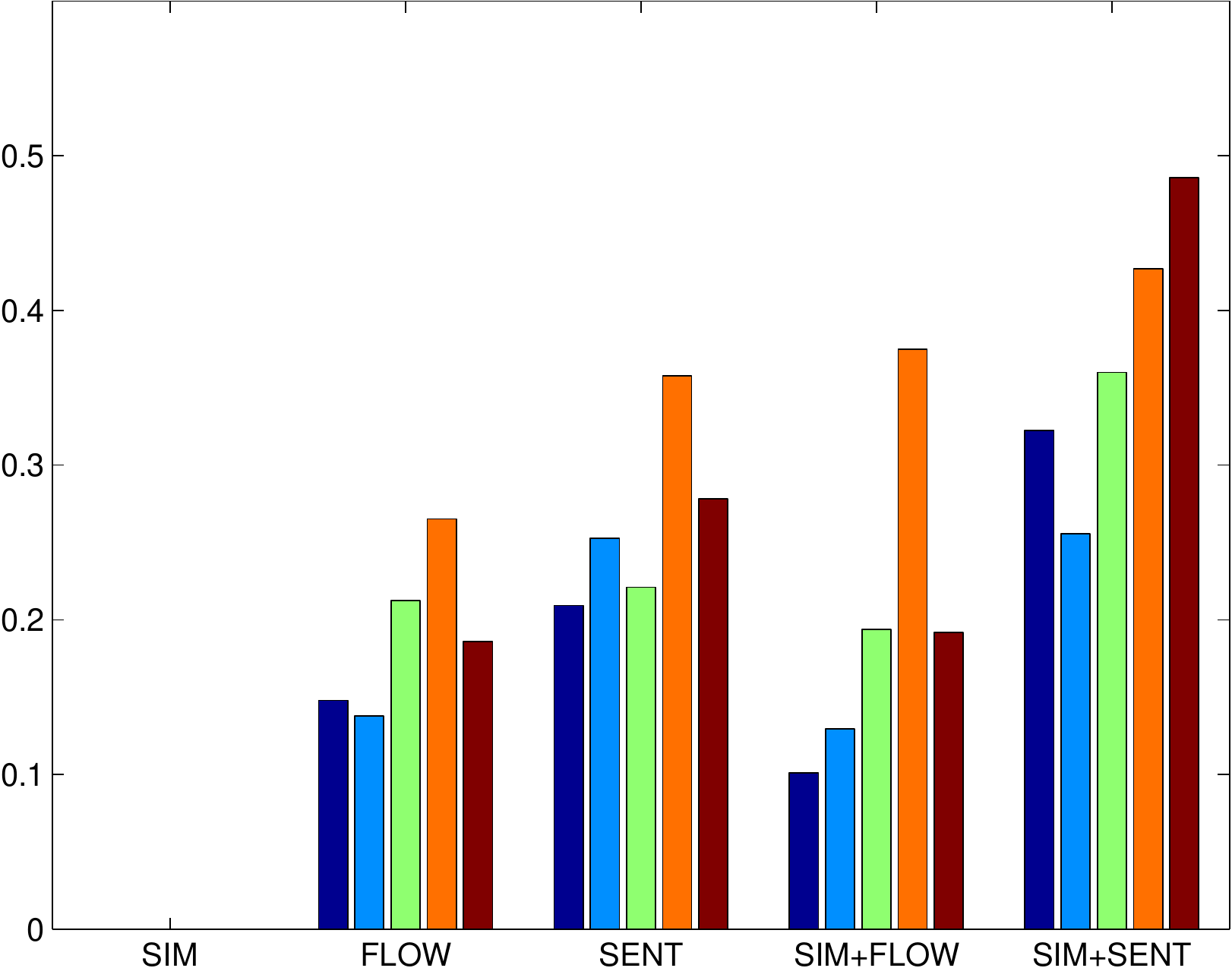}&
      \includegraphics[width=0.33\textwidth]{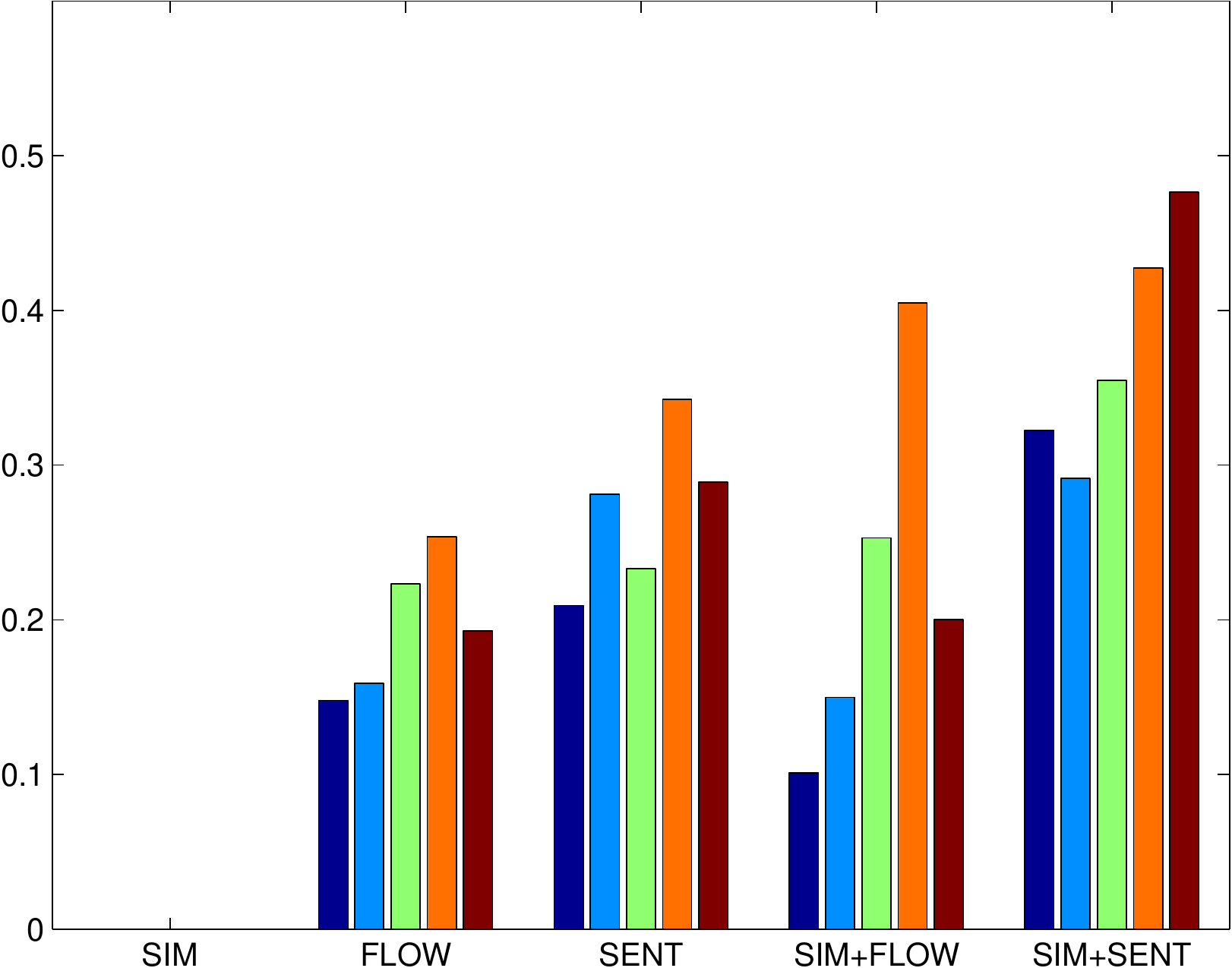}\\
      Run~1&Run~2&Run~3
    \end{tabular}}
  \caption{IoU scores for different variants on different runs of different
    codetection sets on each dataset.
    (top)~Our new dataset, codetection sets
    \usebox{\boxaa}~1,
    \usebox{\boxab}~2,
    \usebox{\boxac}~3,
    \usebox{\boxad}~4,
    \usebox{\boxae}~5,
    \usebox{\boxaf}~6,
    \usebox{\boxag}~7,
    \usebox{\boxah}~8,
    \usebox{\boxai}~9, and
    \usebox{\boxaj}~10.
    (bottom)~Our subset of CAD-120, codetection sets
    \usebox{\boxba}~1,
    \usebox{\boxbb}~2,
    \usebox{\boxbc}~3,
    \usebox{\boxbd}~4, and
    \usebox{\boxbe}~5.}
  \label{fig:iou}
\end{figure*}

\begin{figure*}[t]
  \centering
  \resizebox{\textwidth}{!}{
    \begin{tabular}{ccc}
      \includegraphics[width=0.33\textwidth]{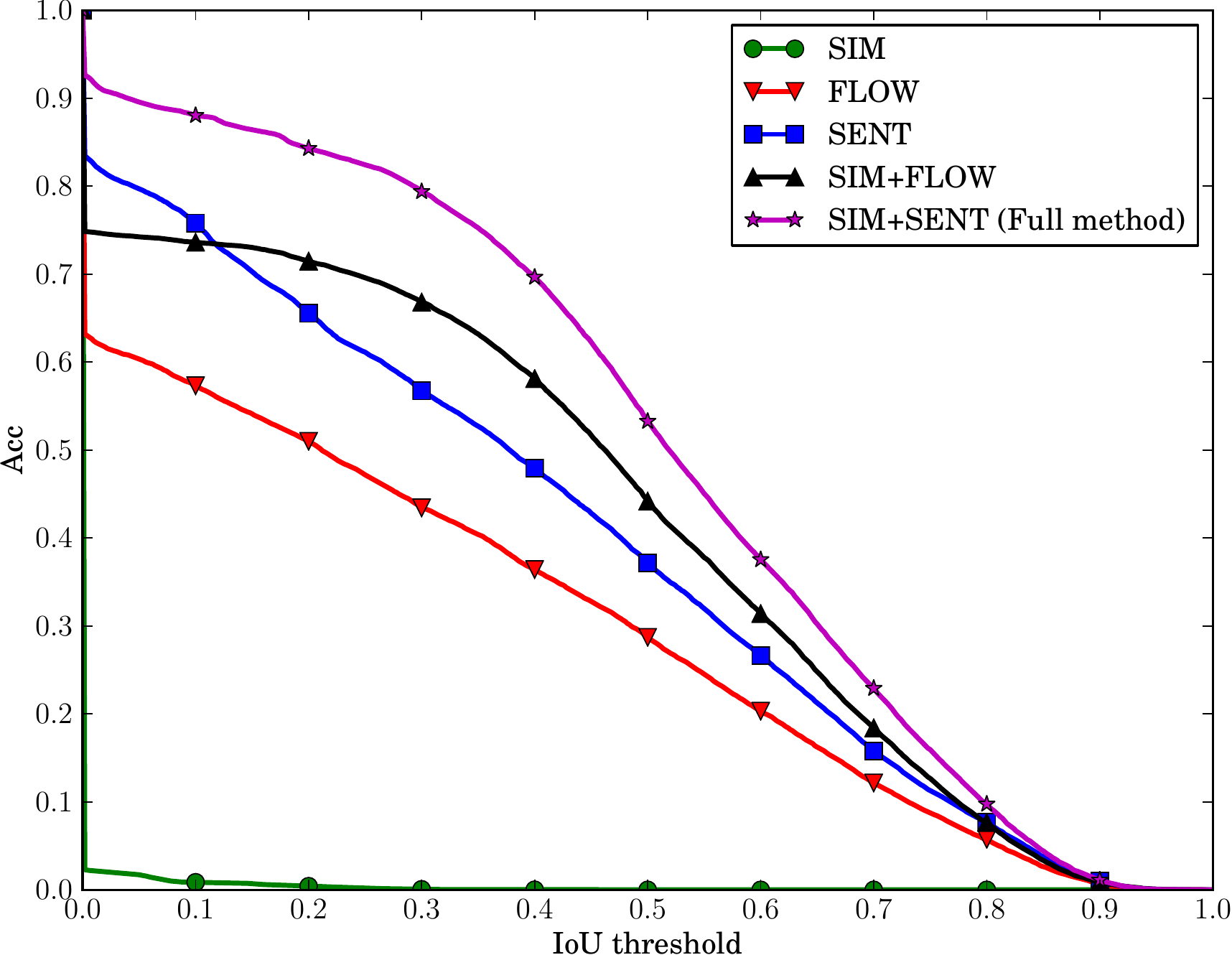}&
      \includegraphics[width=0.33\textwidth]{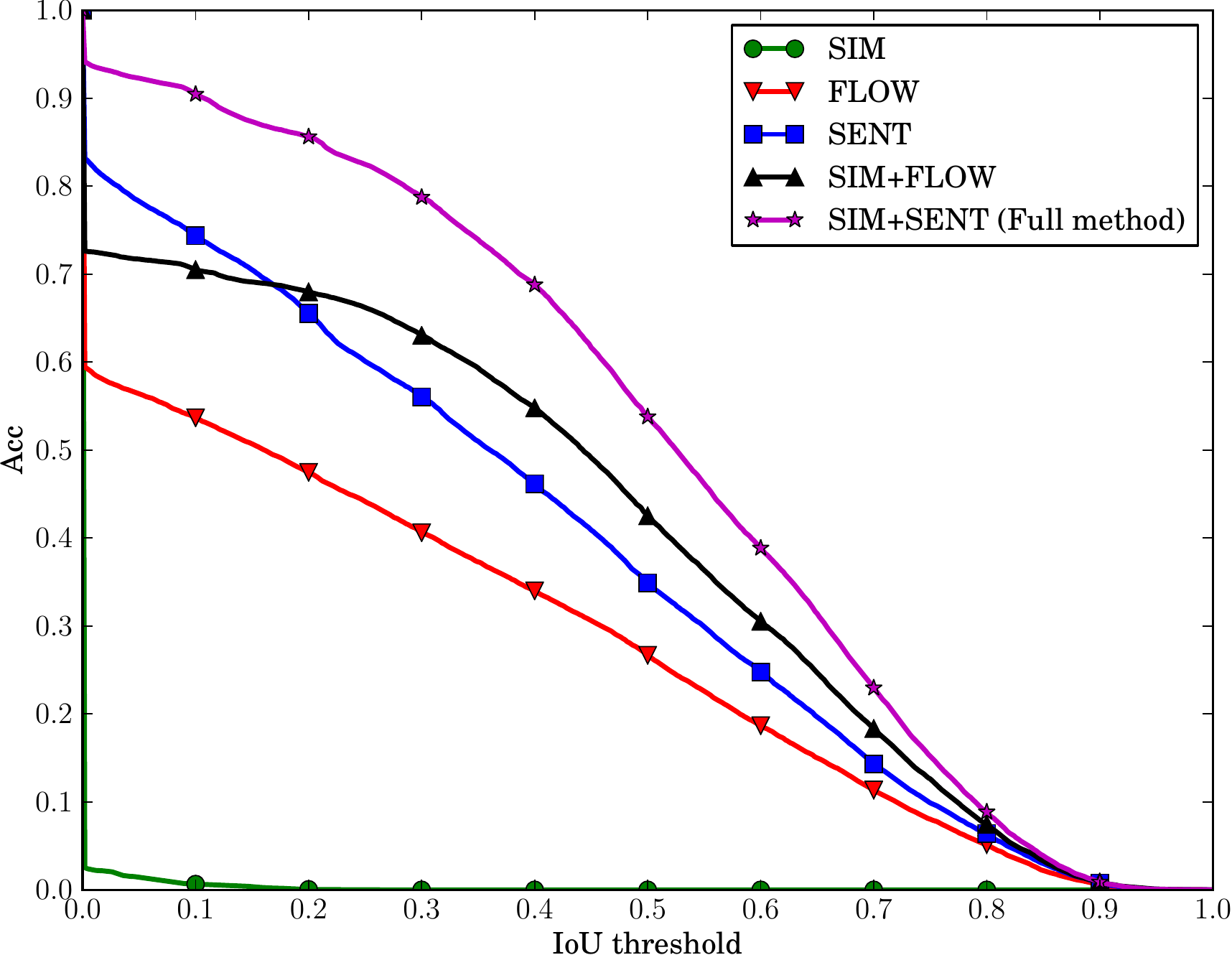}&
      \includegraphics[width=0.33\textwidth]{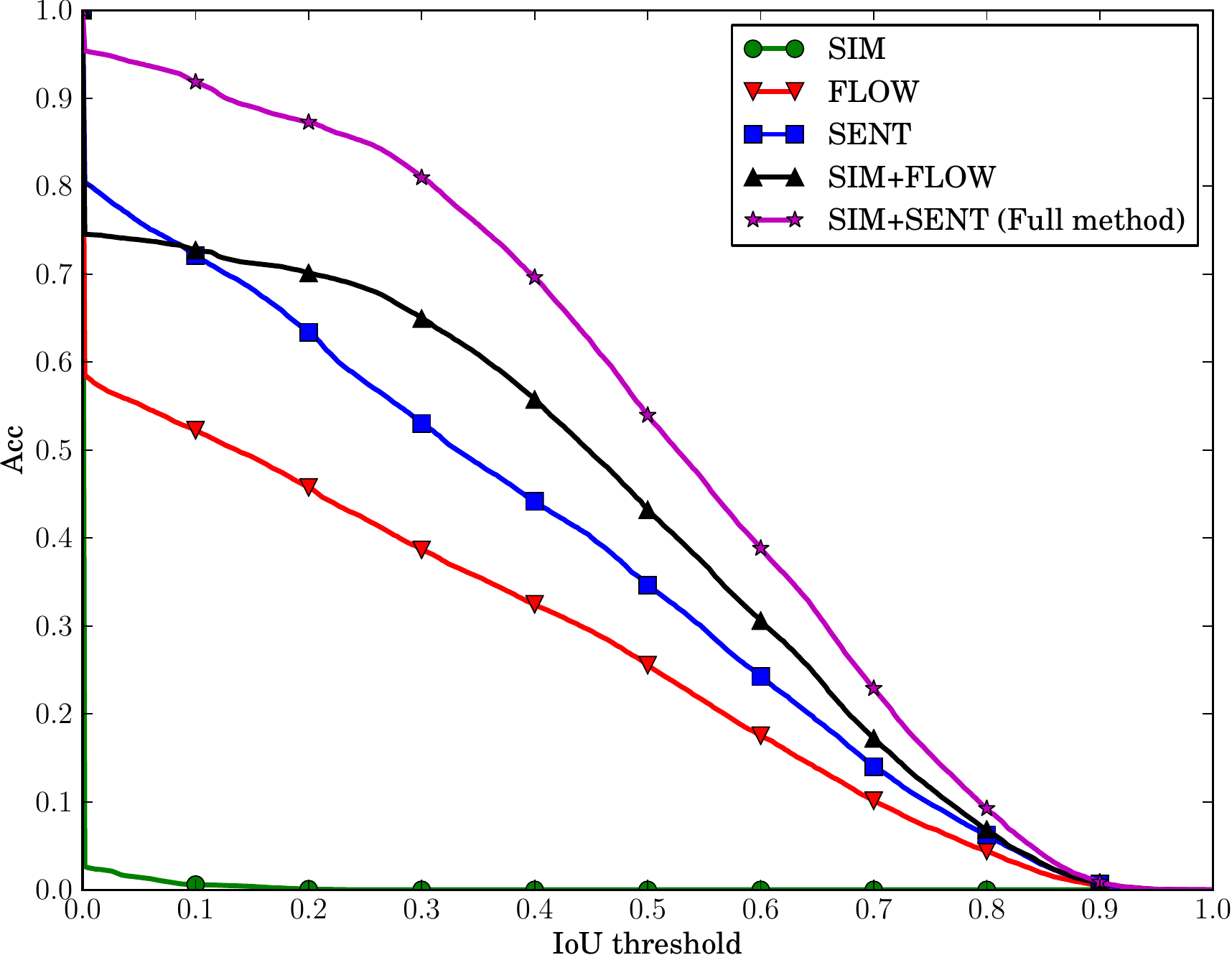}\\
      \includegraphics[width=0.33\textwidth]{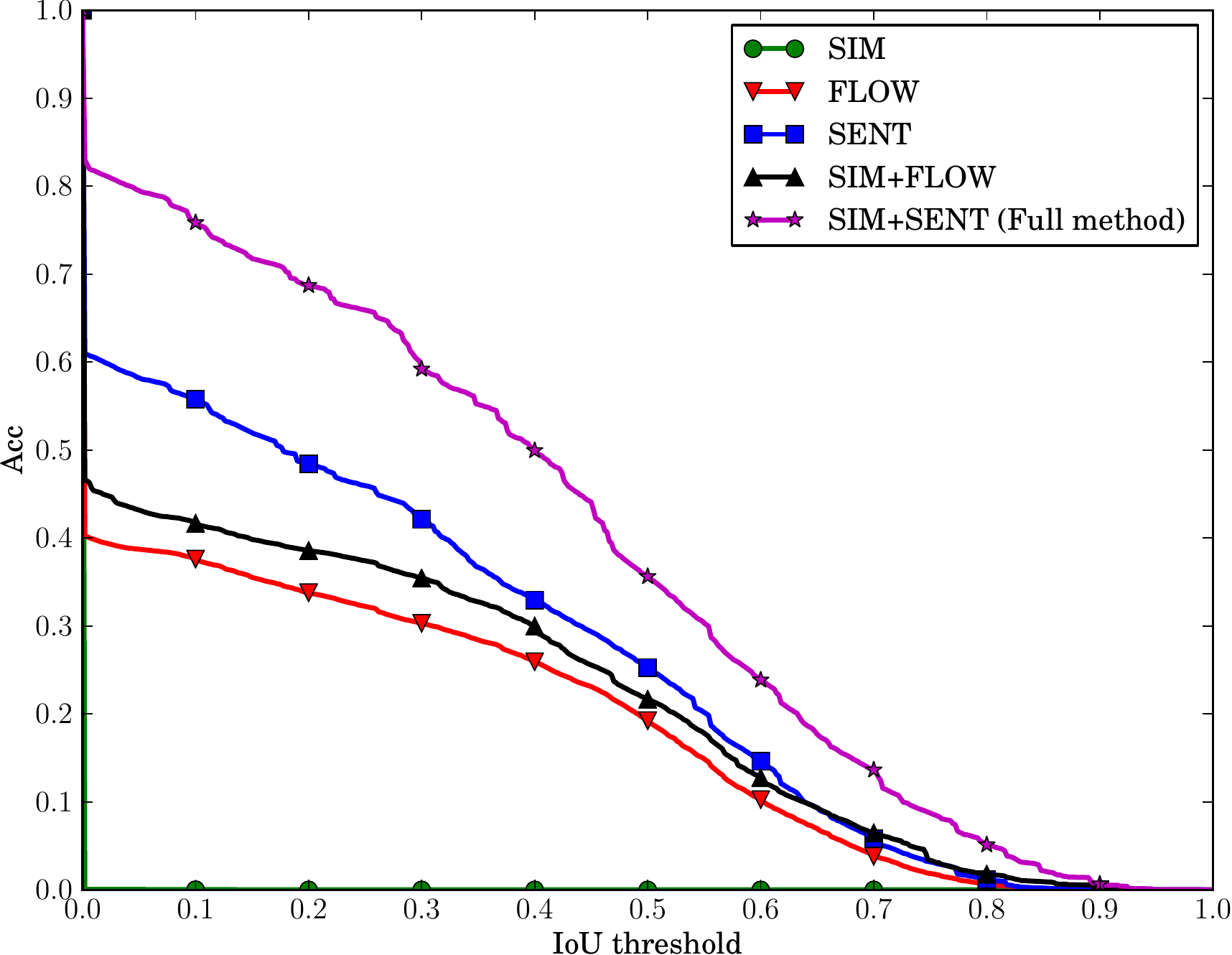}&
      \includegraphics[width=0.33\textwidth]{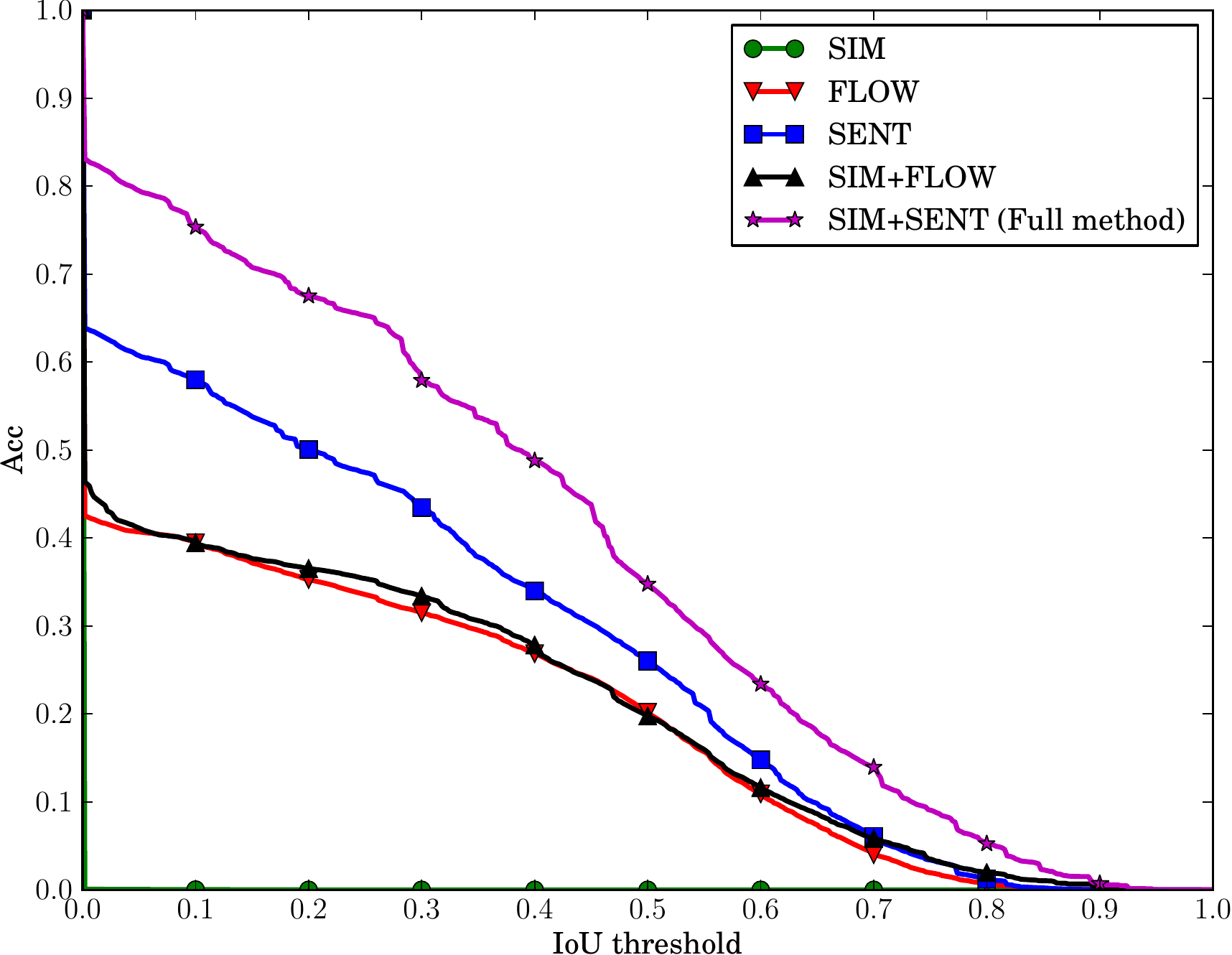}&
      \includegraphics[width=0.33\textwidth]{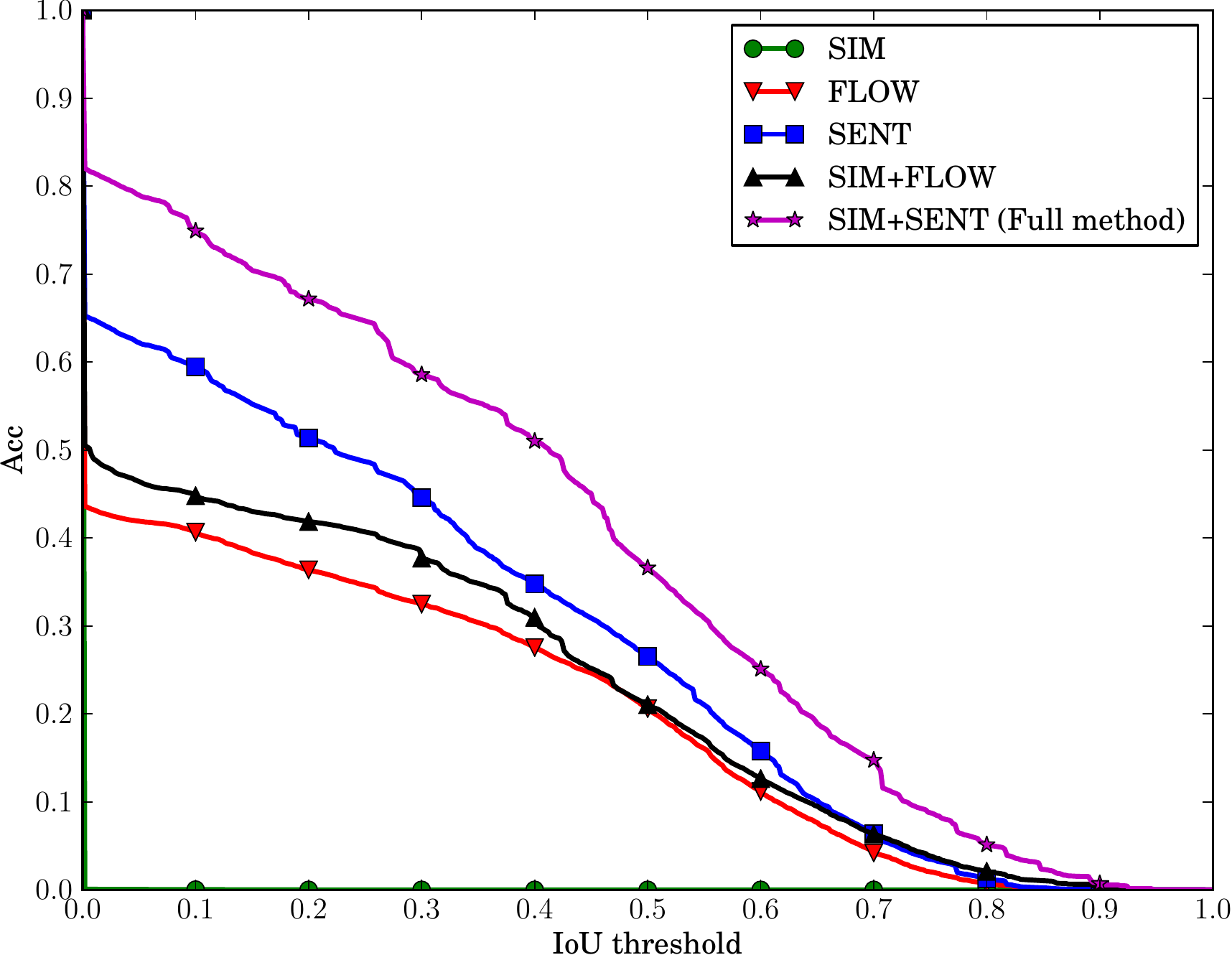}\\
      Run~1&Run~2&Run~3
    \end{tabular}}
  \caption{The codetection accuracy curves of all five methods on our
    datasets.
    (top)~Our new dataset.
    (bottom)~Our subset of CAD-120.}
  \label{fig:acc}
\end{figure*}

Note that \textbf{SIM} uses the similarity measure but no sentential
information.
This method is similar to prior video codetection methods that employ
similarity and the proposal confidence score output by proposal generation
methods to perform codetection.
When the proposal confidence score is not discriminative, as is the case with
our datasets, the prior methods degrade to \textbf{SIM}.
\textbf{FLOW} exploits only binary movement information from the sentence
indicating which objects are probably moving and which are probably not
(\ie\ using only the functions \textsf{medFlMg} and \textsf{tempCoher} in
Table~\ref{tab:predicates}), without similarity or any other sentence semantics
(thus ``partial'' in the table).
\textbf{SIM+FLOW} adds the similarity score on top of \textbf{FLOW}.
\textbf{SENT} uses all possible sentence semantics but no similarity measure.
\textbf{SIM+SENT} is our full method that employs all scores.
All the above variants were applied to each run of each codetection set of
each dataset.
Except for the changes indicated in the above table, all other parameters were
kept constant across all such runs, thus resulting in an apples-to-apples
comparison of the results.
In particular,  $N=500$, $K=240$, $M=20$, and $L=15$ (see
Section~\ref{sec:algorithm} for details).

\subsection{Results}

We quantitatively evaluate our full method and all of the variants by
computing $\text{IoU}_{\text{frame}}$, $\text{IoU}_{\text{object}}$,
$\text{IoU}_{\text{set}}$, and $\text{IoU}_{\text{dataset}}$ for each dataset
as follows.
Given an output box for an object in a video frame, and the
corresponding set of annotated bounding boxes (five boxes for our new dataset
and a single box for CAD-120), we compute IoU scores between the output box
and the annotated ones, and take the averaged IoU score as
$\text{IoU}_{\text{frame}}$.
Then $\text{IoU}_{\text{object}}$ is computed as the average of
$\text{IoU}_{\text{frame}}$ over the output object track.
Then, $\text{IoU}_{\text{set}}$ is computed as the average of
$\text{IoU}_{\text{object}}$ over all the object instances in a codetection
set.
Then, $\text{IoU}_{\text{set}}$ is computed as the average of
$\text{IoU}_{\text{object}}$ over all the object instances in a codetection
set.
Finally $\text{IoU}_{\text{dataset}}$ is computed as the average of
$\text{IoU}_{\text{object}}$ over all runs of all codetection sets for a dataset.

We compute $\text{IoU}_{\text{set}}$ for each variant on each run of each
codetection set in each dataset as shown in Figure~\ref{fig:iou}.
The first variant, \textbf{SIM}, using only the similarity measure,
completely fails on this task as expected.
However, combining \textbf{SIM} with either \textbf{FLOW} or \textbf{SENT}
improves their performance.
Moreover, \textbf{SENT} generally outperforms \textbf{FLOW}, both with and
without the addition of \textbf{SIM}.
Weak information obtained from the sentential annotation that indicates whether
the object is moving or stationary, but no more, \ie\ the distinction between
\textbf{FLOW} and \textbf{SENT}, is helpful in reducing the object proposal
search space, but without the similarity measure, the performance is still
quite poor (\textbf{FLOW}).
Thus one can get moderate results by combining just \textbf{SIM} and
\textbf{FLOW}.
But to further boost performance, more sentence semantics is needed,
\ie\ replacing \textbf{FLOW} with \textbf{SENT}.
Further note that for our new dataset, \textbf{SIM+FLOW} ourperforms
\textbf{SENT}, but for CAD-120, the reverse is true.
This seems to be the case because CAD-120 has greater within-class variance so
sentential information better supports codetection than image similarity.
However, over-constrained semantics can, at times, hinder the codetection
process rather than help, especially given the generality of our datasets.
This is exhibited, for example, with codetection set~4 (\usebox{\boxbd}) on
run~1 of the CAD-120 dataset, where \textbf{SIM+FLOW} outperforms
\textbf{SIM+SENT}.
Thus it is important to only impose \emph{weak} semantics on the codetection
process.

Also note that there is little variation in $\text{IoU}_{\text{set}}$ across
different runs within a dataset.
Recall that the different runs were performed with different sentential
annotations produced by different workers on AMT.\@
This indicates that our approach is largely insensitive to the precise
sentential annotation.

\begin{figure*}[!htbp]
  \setlength{\tabcolsep}{3pt}
  \centering
  \resizebox{0.88\textwidth}{!}{
    \begin{tabular}{cccccccc|cccccccc}
      \hline
      \multicolumn{1}{|c}{\includegraphics[width=0.12\textwidth]{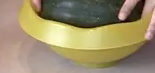}}&
      \includegraphics[width=0.12\textwidth]{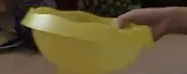}&
      \includegraphics[width=0.12\textwidth]{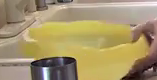}&
      \includegraphics[width=0.12\textwidth]{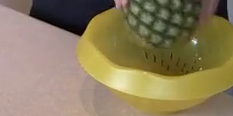}&
      \includegraphics[width=0.12\textwidth]{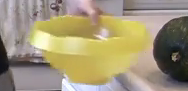}&
      \includegraphics[width=0.12\textwidth]{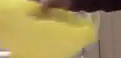}&
      \includegraphics[width=0.12\textwidth]{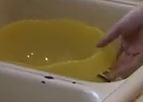}&
      \includegraphics[width=0.12\textwidth]{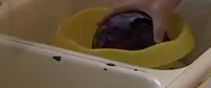}&
      \includegraphics[width=0.12\textwidth]{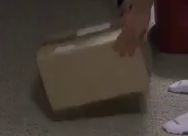}&
      \includegraphics[width=0.12\textwidth]{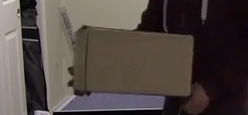}&
      \includegraphics[width=0.12\textwidth]{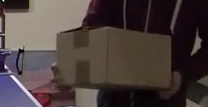}&
      \includegraphics[width=0.12\textwidth]{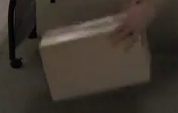}&
      \includegraphics[width=0.12\textwidth]{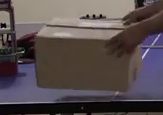}&
      \includegraphics[width=0.12\textwidth]{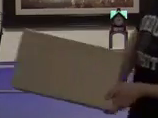}&
      \includegraphics[width=0.12\textwidth]{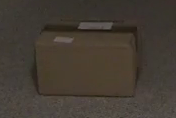}&
      \multicolumn{1}{c|}{\includegraphics[width=0.12\textwidth]{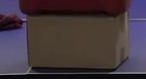}}\\
      \multicolumn{8}{|c|}{\Huge\emph{bowl}}
      &\multicolumn{8}{c|}{\Huge\emph{box}}\\
      \hline
      \multicolumn{1}{|c}{\includegraphics[width=0.12\textwidth]{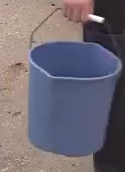}}&
      \includegraphics[width=0.12\textwidth]{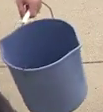}&
      \includegraphics[width=0.12\textwidth]{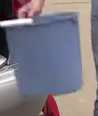}&
      \includegraphics[width=0.12\textwidth]{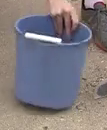}&
      \includegraphics[width=0.12\textwidth]{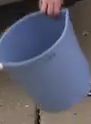}&
      \includegraphics[width=0.12\textwidth]{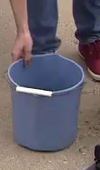}&
      \includegraphics[width=0.12\textwidth]{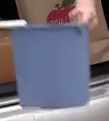}&
      \includegraphics[width=0.12\textwidth]{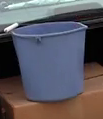}&
      \includegraphics[width=0.12\textwidth]{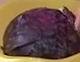}&
      \includegraphics[width=0.12\textwidth]{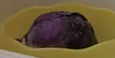}&
      \includegraphics[width=0.12\textwidth]{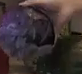}&
      \includegraphics[width=0.12\textwidth]{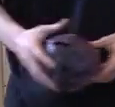}&
      \includegraphics[width=0.12\textwidth]{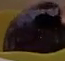}&
      \includegraphics[width=0.12\textwidth]{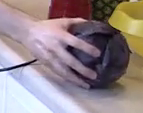}&
      \includegraphics[width=0.12\textwidth]{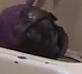}&
      \multicolumn{1}{c|}{\includegraphics[width=0.12\textwidth]{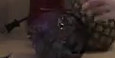}}\\
      \multicolumn{8}{|c|}{\Huge\emph{bucket}}
      &\multicolumn{8}{c|}{\Huge\emph{cabbage}}\\
      \hline
      \multicolumn{1}{|c}{\includegraphics[width=0.12\textwidth]{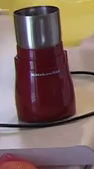}}&
      \includegraphics[width=0.12\textwidth]{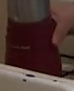}&
      \includegraphics[width=0.12\textwidth]{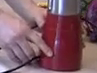}&
      \includegraphics[width=0.12\textwidth]{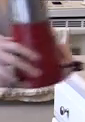}&
      \includegraphics[width=0.12\textwidth]{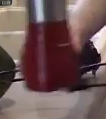}&
      \includegraphics[width=0.12\textwidth]{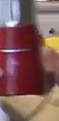}&
      \includegraphics[width=0.12\textwidth]{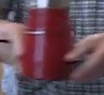}&
      \includegraphics[width=0.12\textwidth]{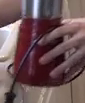}&
      \includegraphics[width=0.12\textwidth]{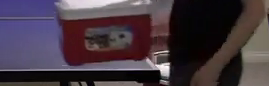}&
      \includegraphics[width=0.12\textwidth]{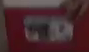}&
      \includegraphics[width=0.12\textwidth]{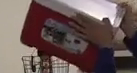}&
      \includegraphics[width=0.12\textwidth]{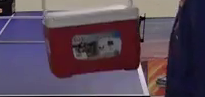}&
      \includegraphics[width=0.12\textwidth]{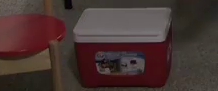}&
      \includegraphics[width=0.12\textwidth]{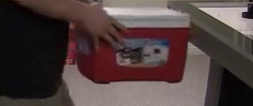}&
      \includegraphics[width=0.12\textwidth]{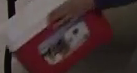}&
      \multicolumn{1}{c|}{\includegraphics[width=0.12\textwidth]{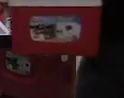}}\\
      \multicolumn{8}{|c|}{\Huge\emph{coffee grinder}}
      &\multicolumn{8}{c|}{\Huge\emph{cooler}}\\
      \hline
      \multicolumn{1}{|c}{\includegraphics[width=0.12\textwidth]{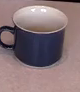}}&
      \includegraphics[width=0.12\textwidth]{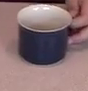}&
      \includegraphics[width=0.12\textwidth]{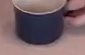}&
      \includegraphics[width=0.12\textwidth]{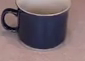}&
      \includegraphics[width=0.12\textwidth]{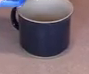}&
      \includegraphics[width=0.12\textwidth]{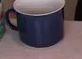}&
      \includegraphics[width=0.12\textwidth]{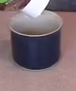}&
      \includegraphics[width=0.12\textwidth]{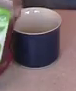}&
      \includegraphics[width=0.12\textwidth]{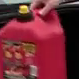}&
      \includegraphics[width=0.12\textwidth]{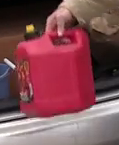}&
      \includegraphics[width=0.12\textwidth]{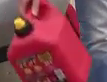}&
      \includegraphics[width=0.12\textwidth]{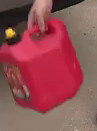}&
      \includegraphics[width=0.12\textwidth]{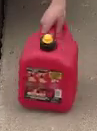}&
      \includegraphics[width=0.12\textwidth]{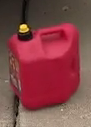}&
      \includegraphics[width=0.12\textwidth]{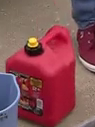}&
      \multicolumn{1}{c|}{\includegraphics[width=0.12\textwidth]{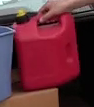}}\\
      \multicolumn{8}{|c|}{\Huge\emph{cup}}
      &\multicolumn{8}{c|}{\Huge\emph{gas can}}\\
      \hline
      \multicolumn{1}{|c}{\includegraphics[width=0.12\textwidth]{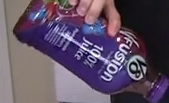}}&
      \includegraphics[width=0.12\textwidth]{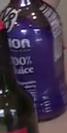}&
      \includegraphics[width=0.12\textwidth]{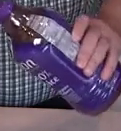}&
      \includegraphics[width=0.12\textwidth]{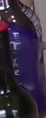}&
      \includegraphics[width=0.12\textwidth]{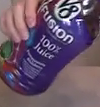}&
      \includegraphics[width=0.12\textwidth]{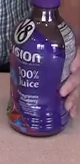}&
      \includegraphics[width=0.12\textwidth]{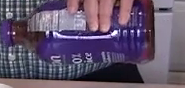}&
      \includegraphics[width=0.12\textwidth]{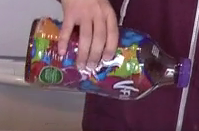}&
      \includegraphics[width=0.12\textwidth]{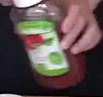}&
      \includegraphics[width=0.12\textwidth]{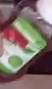}&
      \includegraphics[width=0.12\textwidth]{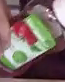}&
      \includegraphics[width=0.12\textwidth]{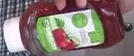}&
      \includegraphics[width=0.12\textwidth]{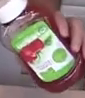}&
      \includegraphics[width=0.12\textwidth]{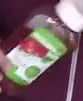}&
      \includegraphics[width=0.12\textwidth]{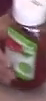}&
      \multicolumn{1}{c|}{\includegraphics[width=0.12\textwidth]{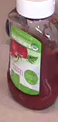}}\\
      \multicolumn{8}{|c|}{\Huge\emph{juice}}
      &\multicolumn{8}{c|}{\Huge\emph{ketchup}}\\
      \hline
      \multicolumn{1}{|c}{\includegraphics[width=0.12\textwidth]{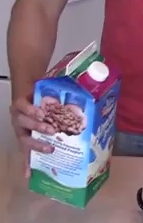}}&
      \includegraphics[width=0.12\textwidth]{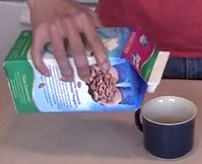}&
      \includegraphics[width=0.12\textwidth]{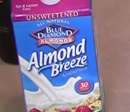}&
      \includegraphics[width=0.12\textwidth]{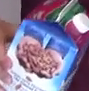}&
      \includegraphics[width=0.12\textwidth]{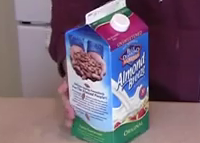}&
      \includegraphics[width=0.12\textwidth]{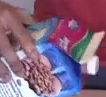}&
      \includegraphics[width=0.12\textwidth]{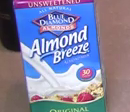}&
      \includegraphics[width=0.12\textwidth]{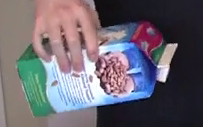}&
      \includegraphics[width=0.12\textwidth]{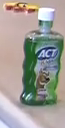}&
      \includegraphics[width=0.12\textwidth]{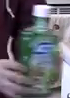}&
      \includegraphics[width=0.12\textwidth]{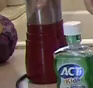}&
      \includegraphics[width=0.12\textwidth]{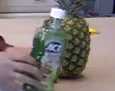}&
      \includegraphics[width=0.12\textwidth]{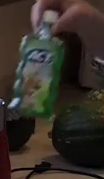}&
      \includegraphics[width=0.12\textwidth]{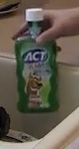}&
      \includegraphics[width=0.12\textwidth]{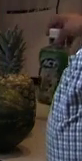}&
      \multicolumn{1}{c|}{\includegraphics[width=0.12\textwidth]{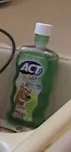}}\\
      \multicolumn{8}{|c|}{\Huge\emph{milk}}
      &\multicolumn{8}{c|}{\Huge\emph{mouthwash}}\\
      \hline
      \multicolumn{1}{|c}{\includegraphics[width=0.12\textwidth]{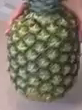}}&
      \includegraphics[width=0.12\textwidth]{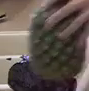}&
      \includegraphics[width=0.12\textwidth]{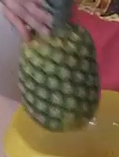}&
      \includegraphics[width=0.12\textwidth]{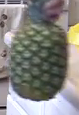}&
      \includegraphics[width=0.12\textwidth]{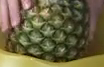}&
      \includegraphics[width=0.12\textwidth]{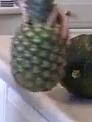}&
      \includegraphics[width=0.12\textwidth]{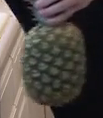}&
      \includegraphics[width=0.12\textwidth]{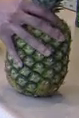}&
      \includegraphics[width=0.12\textwidth]{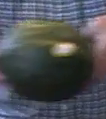}&
      \includegraphics[width=0.12\textwidth]{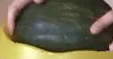}&
      \includegraphics[width=0.12\textwidth]{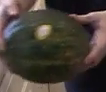}&
      \includegraphics[width=0.12\textwidth]{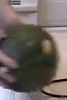}&
      \includegraphics[width=0.12\textwidth]{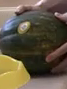}&
      \includegraphics[width=0.12\textwidth]{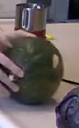}&
      \includegraphics[width=0.12\textwidth]{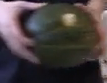}&
      \multicolumn{1}{c|}{\includegraphics[width=0.12\textwidth]{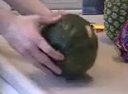}}\\
      \multicolumn{8}{|c|}{\Huge\emph{pineapple}}
      &\multicolumn{8}{c|}{\Huge\emph{squash}}\\
      \hline
      \multicolumn{1}{|c}{\includegraphics[width=0.12\textwidth]{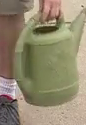}}&
      \includegraphics[width=0.12\textwidth]{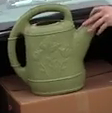}&
      \includegraphics[width=0.12\textwidth]{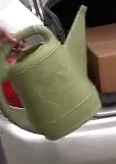}&
      \includegraphics[width=0.12\textwidth]{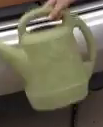}&
      \includegraphics[width=0.12\textwidth]{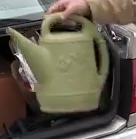}&
      \includegraphics[width=0.12\textwidth]{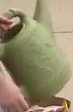}&
      \includegraphics[width=0.12\textwidth]{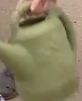}&
      \includegraphics[width=0.12\textwidth]{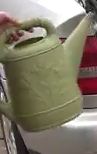}&
      &&&&&&&\\
      \multicolumn{8}{|c|}{\Huge\emph{watering pot}}
      &\multicolumn{8}{c}{}\\
      \cline{1-8}\\
      \hline
      \multicolumn{1}{|c}{\includegraphics[width=0.12\textwidth]{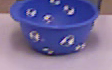}}&
      \includegraphics[width=0.12\textwidth]{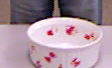}&
      \includegraphics[width=0.12\textwidth]{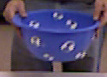}&
      \includegraphics[width=0.12\textwidth]{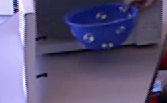}&
      \includegraphics[width=0.12\textwidth]{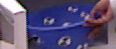}&
      \includegraphics[width=0.12\textwidth]{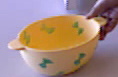}&
      \includegraphics[width=0.12\textwidth]{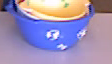}&
      \includegraphics[width=0.12\textwidth]{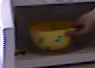}&
      \includegraphics[width=0.12\textwidth]{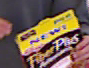}&
      \includegraphics[width=0.12\textwidth]{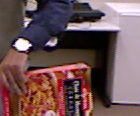}&
      \includegraphics[width=0.12\textwidth]{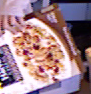}&
      \includegraphics[width=0.12\textwidth]{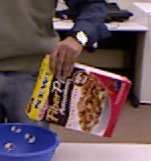}&
      \includegraphics[width=0.12\textwidth]{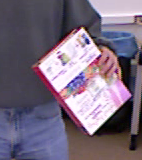}&
      \includegraphics[width=0.12\textwidth]{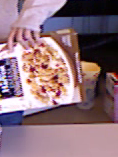}&
      \includegraphics[width=0.12\textwidth]{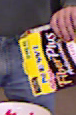}&
      \multicolumn{1}{c|}{\includegraphics[width=0.12\textwidth]{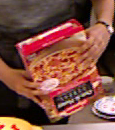}}\\
      \multicolumn{8}{|c|}{\Huge\emph{bowl}}
      &\multicolumn{8}{c|}{\Huge\emph{cereal}}\\
      \hline
      \multicolumn{1}{|c}{\includegraphics[width=0.12\textwidth]{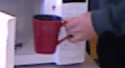}}&
      \includegraphics[width=0.12\textwidth]{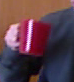}&
      \includegraphics[width=0.12\textwidth]{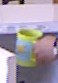}&
      \includegraphics[width=0.12\textwidth]{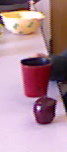}&
      \includegraphics[width=0.12\textwidth]{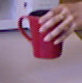}&
      \includegraphics[width=0.12\textwidth]{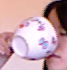}&
      \includegraphics[width=0.12\textwidth]{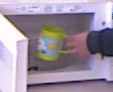}&
      \includegraphics[width=0.12\textwidth]{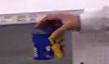}&
      \includegraphics[width=0.12\textwidth]{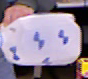}&
      \includegraphics[width=0.12\textwidth]{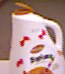}&
      \includegraphics[width=0.12\textwidth]{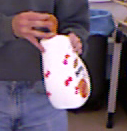}&
      \includegraphics[width=0.12\textwidth]{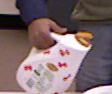}&
      \includegraphics[width=0.12\textwidth]{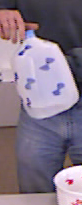}&
      \includegraphics[width=0.12\textwidth]{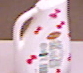}&
      \includegraphics[width=0.12\textwidth]{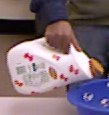}&
      \multicolumn{1}{c|}{\includegraphics[width=0.12\textwidth]{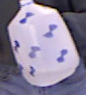}}\\
      \multicolumn{8}{|c|}{\Huge\emph{cup}}
      &\multicolumn{8}{c|}{\Huge\emph{jug}}\\
      \hline
      \multicolumn{1}{|c}{\includegraphics[width=0.12\textwidth]{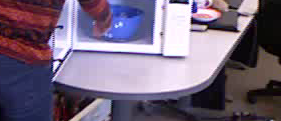}}&
      \includegraphics[width=0.12\textwidth]{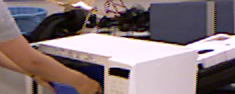}&
      \includegraphics[width=0.12\textwidth]{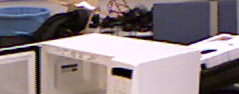}&
      \includegraphics[width=0.12\textwidth]{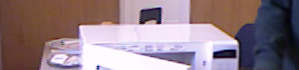}&
      \includegraphics[width=0.12\textwidth]{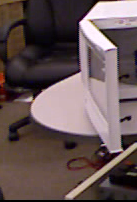}&
      \includegraphics[width=0.12\textwidth]{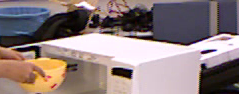}&
      \includegraphics[width=0.12\textwidth]{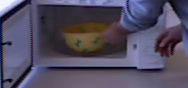}&
      \includegraphics[width=0.12\textwidth]{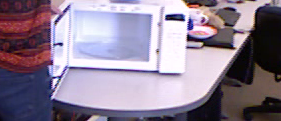}&
      &&&&&&&\\
      \multicolumn{8}{|c|}{\Huge\emph{microwave}}
      &\multicolumn{8}{c}{}\\
      \cline{1-8}
    \end{tabular}}
  \caption{Examples of the 15 codetected object classes in our new dataset
    (top) and the 5 codetected object classes in our subset of CAD-120
    (bottom).
    Note that in some examples the objects are occluded, rotated,
    poorly lit, or blurred due to motion, but they are
    still successfully codetected.
    (For demonstration purposes, the original output detections are
    slightly enlarged to include the surrounding context; zoom in on
    the screen for the best view).}
  \label{fig:examples}
\end{figure*}

To evaluate the performance of our method in simply finding objects, we define
codetection accuracy $\text{Acc}_{\text{frame}}$,
$\text{Acc}_{\text{object}}$, $\text{Acc}_{\text{set}}$, and
$\text{Acc}_{\text{dataset}}$ for each dataset as follows.
Given an IoU threshold, we compute IoU scores between an output box
and the corresponding annotated boxes, and binarize the scores
according to a specified threshold.
Then $\text{Acc}_{\text{frame}}$ is set to the maximum of the
binarized scores, $\text{Acc}_{\text{object}}$ is computed as the
average of $\text{Acc}_{\text{frame}}$ over the output object track,
and $\text{Acc}_{\text{set}}$ is computed as the average of
$\text{Acc}_{\text{object}}$ over all the object instances in a codetection set.
Finally, we average $\text{Acc}_{\text{set}}$ scores over all runs of all
codetection sets for a dataset to obtain $\text{Acc}_{\text{dataset}}$.
By adjusting the IoU threshold from 0 to 1, we get an Acc-vs-threshold
curve for each of the methods (Figure~\ref{fig:acc}).
It can be seen that the codetection accuracies of our full method
under different IoU thresholds consistently outperform those of the
variants.
Our method yields an average detection accuracy
(\ie\ $\text{Acc}_{\text{dataset}}$) of 0.7 to 0.8 on the former (when the IoU
threshold is 0.4 to 0.3) and 0.5 to 0.6 on the latter (when the IoU threshold
is 0.4 to 0.3).
Finally, we demonstrate some codetected object examples in
Figure~\ref{fig:examples}.
For more examples, we refer the readers to our project page.\footnote{\texttt{\url{http://upplysingaoflun.ecn.purdue.edu/~qobi/cccp/sentence-codetection.html}}}

\section{Conclusion}

We have developed a new framework for object codetection in video,
namely, using natural language to guide codetection.
Our experiments indicate that weak sentential information can significantly
improve the results.
This demonstrates that natural language, when combined with typical
computer-vision problems, could provide the capability of high-level reasoning
that yields better solutions to these problems.

\ifCLASSOPTIONcompsoc
  \section*{Acknowledgments}
\else
  \section*{Acknowledgment}
\fi

This research was sponsored, in part, by the Army Research Laboratory,
accomplished under Cooperative Agreement Number W911NF-10-2-0060, and by the
National Science Foundation under Grant No.\ 1522954-IIS.\@
The views, opinions, findings, conclusions, and recommendations contained
in this document are those of the authors and should not be interpreted as
representing the official policies, either express or implied, of the Army
Research Laboratory, the National Science Foundation, or the U.S. Government.
The U.S. Government is authorized to reproduce and distribute reprints for
Government purposes, notwithstanding any copyright notation herein.

\ifCLASSOPTIONcaptionsoff
  \newpage
\fi

\begin{IEEEbiography}
  [{\includegraphics[width=1in,height=1.25in,clip,keepaspectratio]
        {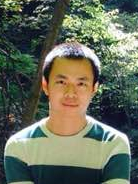}}]{Haonan Yu} is currently a PhD student in the school
    of Electrical and Computer Engineering at Purdue University.
    He received the B.S. degree in Computer Science from Peking University,
    China, in 2011.
    His research interests are Computer Vision and Natural-Language
    Processing, especially combining video and language.
    He was the recipient of the Best Paper Award of ACL 2013.
\end{IEEEbiography}

\begin{IEEEbiography}
  [{\includegraphics[width=1in,height=1.25in,clip,keepaspectratio]
      {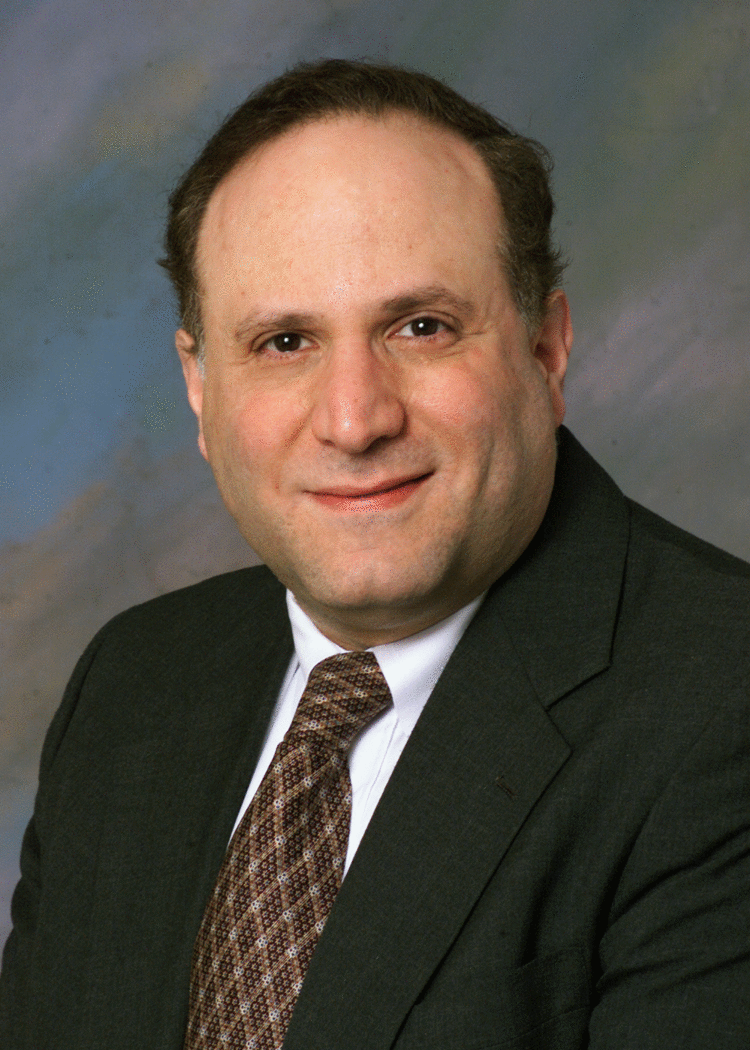}}]{Jeffrey Mark Siskind} received the B.A. degree in
  computer science from the Technion, Israel Institute of Technology in 1979,
  the S.M. degree in computer science from MIT in 1989, and the Ph.D. degree
  in computer science from MIT in 1992.
  He did a postdoctoral fellowship at the University of Pennsylvania
  Institute for Research in Cognitive Science from 1992 to 1993.
  He was an assistant professor at the University of Toronto Department of
  Computer Science from 1993 to 1995, a senior lecturer at the Technion
  Department of Electrical Engineering in 1996, a visiting assistant professor
  at the University of Vermont Department of Computer Science and Electrical
  Engineering from 1996 to 1997, and a research scientist at NEC Research
  Institute, Inc.\ from 1997 to 2001.
  He joined the Purdue University School of Electrical and Computer
  Engineering in 2002 where he is currently an associate professor.
\end{IEEEbiography}


\begin{thebibliography}{46}
\providecommand{\natexlab}[1]{#1}
\providecommand{\url}[1]{\texttt{#1}}
\expandafter\ifx\csname urlstyle\endcsname\relax
  \providecommand{\doi}[1]{doi: #1}\else
  \providecommand{\doi}{doi: \begingroup \urlstyle{rm}\Url}\fi

\bibitem[Alexe et~al.(2010)Alexe, Deselaers, and Ferrari]{Alexe2010}
B.~Alexe, T.~Deselaers, and V.~Ferrari.
\newblock What is an object?
\newblock In \emph{Proceedings of the IEEE Conference on Computer Vision and
  Pattern Recognition}, pages 73--80, 2010.

\bibitem[Andres et~al.(2012)Andres, Beier, and Kappes]{Andres2012}
B.~Andres, T.~Beier, and J.~H. Kappes.
\newblock {OpenGM}: A {C++} library for discrete graphical models.
\newblock \emph{CoRR}, abs/1206.0111, 2012.

\bibitem[Andriluka et~al.(2014)Andriluka, Pishchulin, Gehler, and
  Schiele]{Andriluka2014}
M.~Andriluka, L.~Pishchulin, P.~Gehler, and B.~Schiele.
\newblock {2D} human pose estimation: New benchmark and state of the art
  analysis.
\newblock In \emph{Proceedings of the IEEE Conference on Computer Vision and
  Pattern Recognition}, pages 3686--3693, 2014.

\bibitem[Arbelaez et~al.(2014)Arbelaez, Pont-Tuset, Barron, Marqu{\'{e}}s, and
  Malik]{Arbelaez2014}
P.~Arbelaez, J.~Pont-Tuset, J.~Barron, F.~Marqu{\'{e}}s, and J.~Malik.
\newblock Multiscale combinatorial grouping.
\newblock In \emph{Proceedings of the IEEE Conference on Computer Vision and
  Pattern Recognition}, pages 328--335, 2014.

\bibitem[Barbu et~al.(2012)Barbu, Bridge, Burchill, Coroian, Dickinson, Fidler,
  Michaux, Mussman, Siddharth, Salvi, Schmidt, Shangguan, Siskind, Waggoner,
  Wang, Wei, Yin, and Zhang]{Barbu2012}
A.~Barbu, A.~Bridge, Z.~Burchill, D.~Coroian, S.~Dickinson, S.~Fidler,
  A.~Michaux, S.~Mussman, N.~Siddharth, D.~Salvi, L.~Schmidt, J.~Shangguan,
  J.~M. Siskind, J.~Waggoner, S.~Wang, J.~Wei, Y.~Yin, and Z.~Zhang.
\newblock Video in sentences out.
\newblock In \emph{Proceedings of the Conference on Uncertainty in Artificial
  Intelligence}, pages 102--12, 2012.

\bibitem[Berg et~al.(2004)Berg, Berg, Edwards, Maire, White, Teh,
  Learned-Miller, and Forsyth]{Berg2004}
T.~L. Berg, A.~C. Berg, J.~Edwards, M.~Maire, R.~White, Y.~W. Teh, E.~G.
  Learned-Miller, and D.~A. Forsyth.
\newblock Names and faces in the news.
\newblock In \emph{Proceedings of the IEEE Conference on Computer Vision and
  Pattern Recognition}, pages 848--854, 2004.

\bibitem[Blaschko et~al.(2010)Blaschko, Vedaldi, and Zisserman]{Blaschko2010}
M.~Blaschko, A.~Vedaldi, and A.~Zisserman.
\newblock Simultaneous object detection and ranking with weak supervision.
\newblock In \emph{Advances in Neural Information Processing Systems}, pages
  235--243, 2010.

\bibitem[Bosch et~al.(2007)Bosch, Zisserman, and Munoz]{Bosch2007}
A.~Bosch, A.~Zisserman, and X.~Munoz.
\newblock Image classification using random forests and ferns.
\newblock In \emph{Proceedings of the IEEE International Conference on Computer
  Vision}, pages 1--8, 2007.

\bibitem[Bradski(1998)]{Bradski1998}
G.~R. Bradski.
\newblock Computer vision face tracking for use in a perceptual user interface,
  1998.

\bibitem[Bylinskii et~al.(2012)Bylinskii, Judd, Borji, Itti, Durand, Oliva, and
  Torralba]{Bylinskii2012}
Z.~Bylinskii, T.~Judd, A.~Borji, L.~Itti, F.~Durand, A.~Oliva, and A.~Torralba.
\newblock {MIT} saliency benchmark, 2012.

\bibitem[Cheng et~al.(2014)Cheng, Zhang, Lin, and Torr]{Cheng2014}
M.-M. Cheng, Z.~Zhang, W.-Y. Lin, and P.~Torr.
\newblock {BING}: Binarized normed gradients for objectness estimation at
  300fps.
\newblock In \emph{Proceedings of the IEEE Conference on Computer Vision and
  Pattern Recognition}, pages 3286--3293, 2014.

\bibitem[Clarke et~al.(2010)Clarke, Goldwasser, Chang, and Roth]{Clarke2010}
J.~Clarke, D.~Goldwasser, M.-W. Chang, and D.~Roth.
\newblock Driving semantic parsing from the world's response.
\newblock In \emph{Proceedings of the Conference on Computational Natural
  Language Learning}, pages 18--27, 2010.

\bibitem[Dalal and Triggs(2005)]{Dalal2005}
N.~Dalal and B.~Triggs.
\newblock Histograms of oriented gradients for human detection.
\newblock In \emph{Proceedings of the IEEE Conference on Computer Vision and
  Pattern Recognition}, pages 886--893, 2005.

\bibitem[Das et~al.(2013)Das, Xu, Doell, and Corso]{Das2013}
P.~Das, C.~Xu, R.~F. Doell, and J.~J. Corso.
\newblock A thousand frames in just a few words: Lingual description of videos
  through latent topics and sparse object stitching.
\newblock In \emph{Proceedings of the IEEE Conference on Computer Vision and
  Pattern Recognition}, pages 2634--2641, 2013.

\bibitem[Farneb{\"{a}}ck(2003)]{Farneback2003}
G.~Farneb{\"{a}}ck.
\newblock Two-frame motion estimation based on polynomial expansion.
\newblock In \emph{Proceedings of the Scandinavian Conference on Image
  Analysis}, pages 363--370, 2003.

\bibitem[Fossati et~al.(2013)Fossati, Gall, Grabner, Ren, and
  Konolige]{Fossati2013}
A.~Fossati, J.~Gall, H.~Grabner, X.~Ren, and K.~Konolige.
\newblock \emph{Consumer Depth Cameras for Computer Vision}, chapter Human Body
  Analysis.
\newblock Springer, 2013.

\bibitem[Guadarrama et~al.(2013)Guadarrama, Krishnamoorthy, Malkarnenkar,
  Venugopalan, Mooney, Darrell, and Saenko]{Guadarrama2013}
S.~Guadarrama, N.~Krishnamoorthy, G.~Malkarnenkar, S.~Venugopalan, R.~Mooney,
  T.~Darrell, and K.~Saenko.
\newblock {YouTube2Text}: Recognizing and describing arbitrary activities using
  semantic hierarchies and zero-shot recognition.
\newblock In \emph{Proceedings of the IEEE International Conference on Computer
  Vision}, pages 2712--2719, 2013.

\bibitem[Gupta and Davis(2008)]{Gupta2008}
A.~Gupta and L.~S. Davis.
\newblock Beyond nouns: Exploiting prepositions and comparative adjectives for
  learning visual classifiers.
\newblock In \emph{Proceedings of the European Conference on Computer Vision},
  pages 16--29, 2008.

\bibitem[Jamieson et~al.(2010{\natexlab{a}})Jamieson, Eskin, Fazly, Stevenson,
  and Dickinson]{Jamieson2010b}
M.~Jamieson, Y.~Eskin, A.~Fazly, S.~Stevenson, and S.~Dickinson.
\newblock Discovering multipart appearance models from captioned images.
\newblock In \emph{Proceedings of the European Conference on Computer Vision},
  pages 183--196, 2010{\natexlab{a}}.

\bibitem[Jamieson et~al.(2010{\natexlab{b}})Jamieson, Fazly, Stevenson,
  Dickinson, and Wachsmuth]{Jamieson2010a}
M.~Jamieson, A.~Fazly, S.~Stevenson, S.~J. Dickinson, and S.~Wachsmuth.
\newblock Using language to learn structured appearance models for image
  annotation.
\newblock \emph{IEEE Transactions on Pattern Analysis and Machine
  Intelligence}, 32\penalty0 (1):\penalty0 148--164, 2010{\natexlab{b}}.

\bibitem[Jiang et~al.(2015)Jiang, Huang, Duan, and Zhao]{Jiang2015}
M.~Jiang, S.~Huang, J.~Duan, and Q.~Zhao.
\newblock {SALICON}: Saliency in context.
\newblock In \emph{Proceedings of the IEEE Conference on Computer Vision and
  Pattern Recognition}, 2015.

\bibitem[Joulin et~al.(2014)Joulin, Tang, and Fei-Fei]{Armand2014}
A.~Joulin, K.~Tang, and L.~Fei-Fei.
\newblock Efficient image and video co-localization with {F}rank-{W}olfe
  algorithm.
\newblock In \emph{Proceedings of the European Conference on Computer Vision},
  pages 253--268, 2014.

\bibitem[Kong et~al.(2014)Kong, Lin, Bansal, Urtasun, and Fidler]{Kong2014}
C.~Kong, D.~Lin, M.~Bansal, R.~Urtasun, and S.~Fidler.
\newblock What are you talking about? text-to-image coreference.
\newblock In \emph{Proceedings of the IEEE Conference on Computer Vision and
  Pattern Recognition}, pages 3558--3565, 2014.

\bibitem[Koppula et~al.(2013)Koppula, Gupta, and Saxena]{Koppula2013}
H.~S. Koppula, R.~Gupta, and A.~Saxena.
\newblock Learning human activities and object affordances from {RGB-D} videos.
\newblock \emph{International Journal of Robotics Research}, 32\penalty0
  (8):\penalty0 951--970, 2013.

\bibitem[Lee and Grauman(2011)]{Lee2011}
Y.~J. Lee and K.~Grauman.
\newblock Learning the easy things first: Self-paced visual category discovery.
\newblock In \emph{Proceedings of the IEEE Conference on Computer Vision and
  Pattern Recognition}, pages 1721--1728, 2011.

\bibitem[Lin et~al.(2014)Lin, Fidler, Kong, and Urtasun]{Lin2014}
D.~Lin, S.~Fidler, C.~Kong, and R.~Urtasun.
\newblock Visual semantic search: Retrieving videos via complex textual
  queries.
\newblock In \emph{Proceedings of the IEEE Conference on Computer Vision and
  Pattern Recognition}, pages 2657--2664, 2014.

\bibitem[Lowe(2004)]{Lowe2004}
D.~G. Lowe.
\newblock Distinctive image features from scale-invariant keypoints.
\newblock \emph{International Journal of Computer Vision}, 60\penalty0
  (2):\penalty0 91--110, 2004.

\bibitem[Luo et~al.(2009)Luo, Caputo, and Ferrari]{Luo2009}
J.~Luo, B.~Caputo, and V.~Ferrari.
\newblock Who’s doing what: Joint modeling of names and verbs for
  simultaneous face and pose annotation.
\newblock In \emph{Advances in Neural Information Processing Systems}, pages
  1168--1176, 2009.

\bibitem[Marsza{\l}ek et~al.(2009)Marsza{\l}ek, Laptev, and
  Schmid]{Marszalek2009}
M.~Marsza{\l}ek, I.~Laptev, and C.~Schmid.
\newblock Actions in context.
\newblock In \emph{Proceedings of the IEEE Conference on Computer Vision and
  Pattern Recognition}, pages 2929--2936, 2009.

\bibitem[Moringen et~al.(2008)Moringen, Wachsmuth, Dickinson, and
  Stevenson]{Moringen2008}
J.~Moringen, S.~Wachsmuth, S.~Dickinson, and S.~Stevenson.
\newblock Learning visual compound models from parallel image-text datasets.
\newblock \emph{Pattern Recognition}, 5096:\penalty0 486--496, 2008.

\bibitem[Palmer et~al.(2005)Palmer, Gildea, and Kingsbury]{Palmer2005}
M.~Palmer, D.~Gildea, and P.~Kingsbury.
\newblock The proposition bank: An annotated corpus of semantic roles.
\newblock \emph{Computational linguistics}, 31\penalty0 (1):\penalty0 71--106,
  2005.

\bibitem[Pearl(1982)]{Pearl1982}
J.~Pearl.
\newblock Reverend {B}ayes on inference engines: a distributed hierarchical
  approach.
\newblock In \emph{Proceedings of the Conference on Artificial Intelligence},
  pages 133--136, 1982.

\bibitem[Plummer et~al.(2015)Plummer, Wang, Cervantes, Caicedo, Hockenmaier,
  and Lazebnik]{Plummer2015}
B.~Plummer, L.~Wang, C.~Cervantes, J.~Caicedo, J.~Hockenmaier, and S.~Lazebnik.
\newblock Flickr30k entities: Collecting region-to-phrase correspondences for
  richer image-to-sentence models.
\newblock In \emph{Proceedings of the IEEE International Conference on Computer
  Vision}, 2015.

\bibitem[Prest et~al.(2012)Prest, Leistner, Civera, Schmid, and
  Ferrari]{Prest2012}
A.~Prest, C.~Leistner, J.~Civera, C.~Schmid, and V.~Ferrari.
\newblock Learning object class detectors from weakly annotated video.
\newblock In \emph{Proceedings of the IEEE Conference on Computer Vision and
  Pattern Recognition}, pages 3282--3289, 2012.

\bibitem[Ramanathan et~al.(2014)Ramanathan, Joulin, Liang, and
  Fei-Fei]{Ramanathan2014}
V.~Ramanathan, A.~Joulin, P.~Liang, and L.~Fei-Fei.
\newblock Linking people with ``their'' names using coreference resolution.
\newblock In \emph{Proceedings of the European Conference on Computer Vision},
  pages 95--110, 2014.

\bibitem[Rodriguez et~al.(2008)Rodriguez, Ahmed, and Shah]{Rodriguez2008}
M.~D. Rodriguez, J.~Ahmed, and M.~Shah.
\newblock Action {MACH}: A spatio-temporal maximum average correlation height
  filter for action recognition.
\newblock In \emph{Proceedings of the IEEE Conference on Computer Vision and
  Pattern Recognition}, pages 1--8, 2008.

\bibitem[Rohrbach et~al.(2015)Rohrbach, Rohrbach, Tandon, and
  Schiele]{Rohrbach2015}
A.~Rohrbach, M.~Rohrbach, N.~Tandon, and B.~Schiele.
\newblock A dataset for movie description.
\newblock In \emph{Proceedings of the IEEE Conference on Computer Vision and
  Pattern Recognition}, 2015.

\bibitem[Rohrbach et~al.(2012)Rohrbach, Amin, Andriluka, and
  Schiele]{Rohrbach2012}
M.~Rohrbach, S.~Amin, M.~Andriluka, and B.~Schiele.
\newblock A database for fine grained activity detection of cooking activities.
\newblock In \emph{Proceedings of the IEEE Conference on Computer Vision and
  Pattern Recognition}, pages 1194--1201, 2012.

\bibitem[Rubinstein et~al.(2013)Rubinstein, Joulin, Kopf, and
  Liu]{Rubinstein2013}
M.~Rubinstein, A.~Joulin, J.~Kopf, and C.~Liu.
\newblock Unsupervised joint object discovery and segmentation in internet
  images.
\newblock In \emph{Proceedings of the IEEE Conference on Computer Vision and
  Pattern Recognition}, pages 1939--1946, 2013.

\bibitem[Schulter et~al.(2013)Schulter, Leistner, Roth, and
  Bischof]{Schulter2013}
S.~Schulter, C.~Leistner, P.~M. Roth, and H.~Bischof.
\newblock Unsupervised object discovery and segmentation in videos.
\newblock In \emph{Proceedings of the British Machine Vision Conference}, pages
  53.1--53.12, 2013.

\bibitem[Srikantha and Gall(2014)]{Srikantha2014}
A.~Srikantha and J.~Gall.
\newblock Discovering object classes from activities.
\newblock In \emph{Proceedings of the European Conference on Computer Vision},
  pages 415--430, 2014.

\bibitem[Tang et~al.(2014)Tang, Joulin, Li, and Fei-Fei]{Tang2014}
K.~Tang, A.~Joulin, J.~Li, and L.~Fei-Fei.
\newblock Co-localization in real-world images.
\newblock In \emph{Proceedings of the IEEE Conference on Computer Vision and
  Pattern Recognition}, pages 1464--1471, 2014.

\bibitem[Torabi et~al.(2015)Torabi, Chris, Hugo, and Aaron]{Torabi2015}
A.~Torabi, P.~Chris, L.~Hugo, and C.~Aaron.
\newblock Using descriptive video services to create a large data source for
  video annotation research.
\newblock \emph{CoRR}, abs/1503.01070, 2015.

\bibitem[Tuytelaars et~al.(2010)Tuytelaars, Lampert, Blaschko, and
  Buntine]{Tuytelaars2010}
T.~Tuytelaars, C.~H. Lampert, M.~B. Blaschko, and W.~L. Buntine.
\newblock Unsupervised object discovery: A comparison.
\newblock \emph{International Journal of Computer Vision}, 88\penalty0
  (2):\penalty0 284--302, 2010.

\bibitem[Wong and Mooney(2007)]{Wong2007}
Y.~W. Wong and R.~J. Mooney.
\newblock Learning synchronous grammars for semantic parsing with lambda
  calculus.
\newblock In \emph{Proceedings of the Annual Meeting of the Association for
  Computational Linguistics}, pages 960--967, 2007.

\bibitem[Zitnick and Doll{\'{a}}r(2014)]{Zitnick2014}
C.~L. Zitnick and P.~Doll{\'{a}}r.
\newblock Edge boxes: Locating object proposals from edges.
\newblock In \emph{Proceedings of the European Conference on Computer Vision},
  pages 391--405, 2014.

\end{thebibliography}
\end{document}